\documentclass[runningheads]{llncs}

\usepackage{eccv}

\usepackage{algorithm,algorithmic}
\usepackage{array}
\usepackage{amsfonts}
\usepackage{amsmath}
\usepackage{mathalfa}
\usepackage{makecell}
\usepackage{breqn}
\usepackage[bottom]{footmisc}

\usepackage{xcolor}
\usepackage{listings}
\usepackage{color}

\usepackage{paralist}

\usepackage{graphicx}
\usepackage{booktabs} 
\newcommand*{\img}[1]{%
    \raisebox{-.03\baselineskip}{%
        \includegraphics[
        height=\baselineskip,
        width=\baselineskip,
        keepaspectratio,
        ]{#1}%
    }%
}
\usepackage{bbding}

\usepackage{textcomp,booktabs}
\usepackage{colortbl}
\definecolor{mygray}{gray}{.9}
\definecolor{mypink}{rgb}{.99,.91,.95}
\definecolor{mycyan}{cmyk}{.3,0,0,0}
\definecolor{textred}{RGB}{234,56,28}
\definecolor{textgreen}{RGB}{94,128,59}
\definecolor{textorange}{RGB}{233,187,150}
\usepackage{eccvabbrv}

\usepackage{graphicx}
\usepackage{booktabs}

\usepackage[accsupp]{axessibility}  % Improves PDF readability for those with disabilities.
\usepackage{xcolor,pifont}
\usepackage{fontawesome5}
\usepackage{xcolor,pifont}
\newcommand*\colourcheck[1]{%
  \expandafter\newcommand\csname #1check\endcsname{\textcolor{#1}{\ding{52}}}%
}
\newcommand*\colourcross[1]{%
  \expandafter\newcommand\csname #1check\endcsname{\textcolor{#1}{\ding{55}}}%
}

\colourcheck{green}

\colourcross{red}
\newcommand{\NAME}{\textsc{FineMatch}\xspace}
\newcommand{\AT}{\textsc{AutoAlign}\xspace}
\usepackage[pagebackref,breaklinks,colorlinks,citecolor=eccvblue]{hyperref}
\usepackage{hyperref}

\usepackage{orcidlink}

\begin{document}

\title{\img{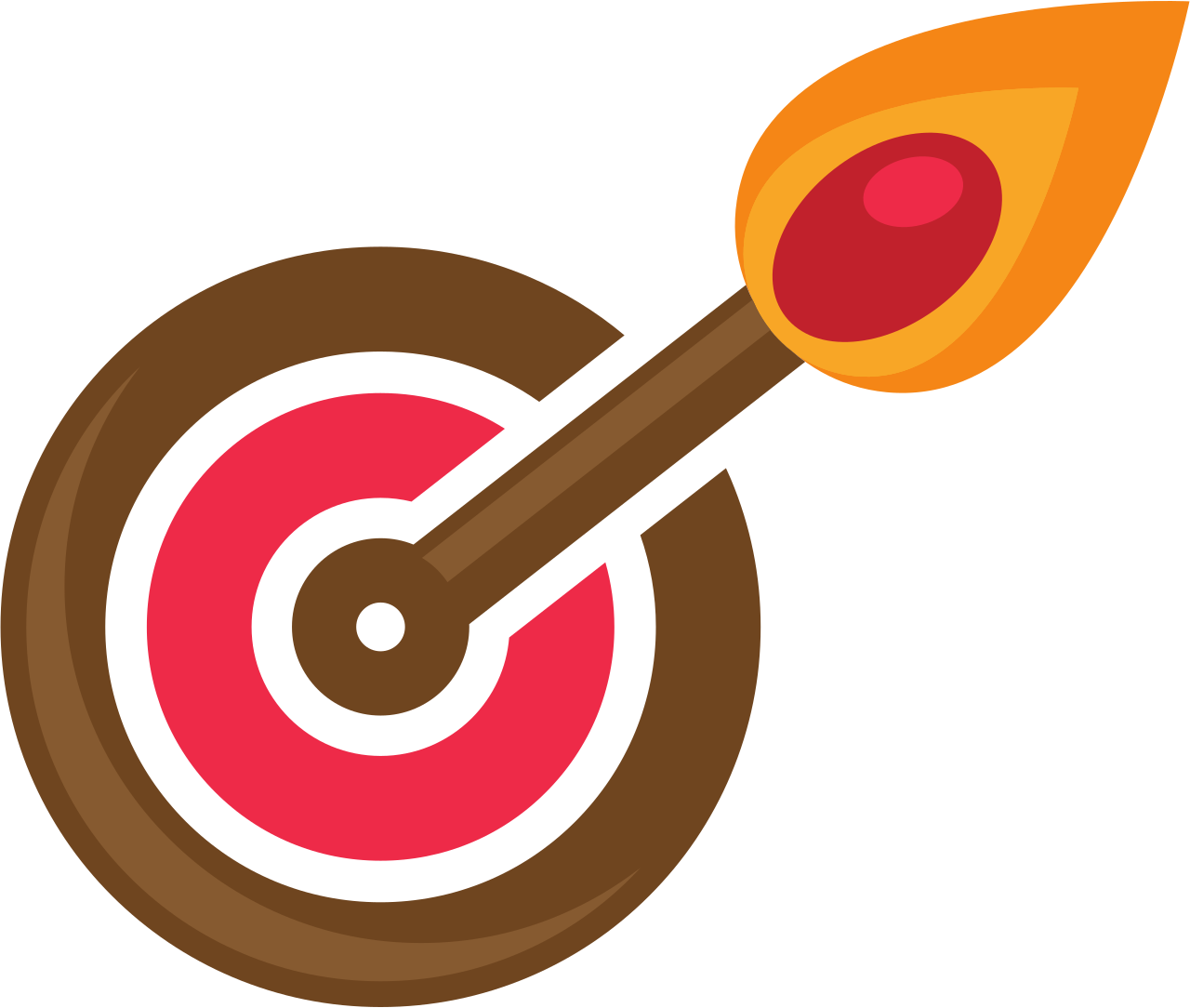} \NAME: Aspect-based Fine-grained Image and Text Mismatch Detection and Correction} 
%\img{img/Icon3.png} 
\titlerunning{\NAME}

%\author{First Author\inst{1}\orcidlink{0000-1111-2222-3333} \and
%Second Author\inst{2,3}\orcidlink{1111-2222-3333-4444} \and
%Third Author\inst{3}\orcidlink{2222--3333-4444-5555}}

\author{
  Hang Hua \textsuperscript{1}\and 
  Jing Shi \textsuperscript{2}\and
  Kushal Kafle\textsuperscript{2}\and
  Simon Jenni\textsuperscript{2}\and
  Daoan Zhang\textsuperscript{1}\and \\
  John Collomosse\textsuperscript{2}\and
  Scott Cohen\textsuperscript{2}\and
  Jiebo Luo\textsuperscript{1}
}
\authorrunning{H.~Hua et al.}

\institute{University of Rochester \and
Adobe Research \\
\email{\{hhua2, jluo\}@cs.rochester.edu, dzhang52@ur.rochester.edu}\\
\email{\{jingshi, kkafle, jenni, collomos, scohen\}@adobe.com}}
\maketitle

\begin{abstract}
   Recent progress in large-scale pre-training has led to the development of advanced vision-language models (VLMs) with remarkable proficiency in comprehending and generating multimodal content. Despite the impressive ability to perform complex reasoning for VLMs, current models often struggle to effectively and precisely capture the compositional information on both the image and text sides. To address this, we propose \textbf{\NAME}, a new aspect-based fine-grained text and image matching benchmark, focusing on text and image mismatch detection and correction. This benchmark introduces a novel task for boosting and evaluating the VLMs' compositionality for aspect-based fine-grained text and image matching. In this task, models are required to identify mismatched aspect phrases within a caption, determine the aspect's class, and propose corrections for an image-text pair that may contain between 0 and 3 mismatches. To evaluate the models' performance on this new task, we propose a new evaluation metric named \textbf{ITM-IoU} for which our experiments show a high correlation to human evaluation. In addition, we also provide a comprehensive experimental analysis of existing mainstream VLMs, including fully supervised learning and in-context learning settings. We have found that models trained on \NAME demonstrate enhanced proficiency in detecting fine-grained text and image mismatches. Moreover, models (e.g., GPT-4V, Gemini Pro Vision) with strong abilities to perform multimodal in-context learning are not as skilled at fine-grained compositional image and text matching analysis. With \NAME, we are able to build a system for text-to-image generation hallucination detection and correction. Resources are available at \url{https://hanghuacs.github.io/finematch/}.
  \keywords{Pre-trained Vision-Language Models \and Aspect-based Image and Text Analysis \and Compositionality}
\end{abstract}
\section{Introduction}
\label{sec:intro}
Pretrained vision-language models, such as GPT-4V \cite{2023GPT4VisionSC}, LLaVA \cite{Touvron2023LLaMAOA}, MiniGPT-4 \cite{chen2023minigptv2}, and BLIP \cite{li2022blip,li2023blip},
have demonstrated impressive ability to perform complex reasoning. Benefiting from large pretrained VLMs, a series of VLM-based methods \cite{hu2022promptcap,lin2023videoxum,tang2024avicuna,hua2024v2xum,yu2024promptfix} have emerged and achieved remarkable results on various vision-language (VL) tasks. However, contemporary state-of-the-art VLMs still struggle with fine-grained compositional information understanding \cite{yuksekgonul2022and, hudson2018gqa}. Prior works have pointed out that pretrained VLMs face challenges in comprehending fine-grained visual and textual compositional information \cite{diwan2022winoground,yuksekgonul2022and}. This issue poses a significant limitation to the reliability and performance of VLMs.

\begin{figure}[t!]
\centering
  \includegraphics[width=0.9\textwidth]{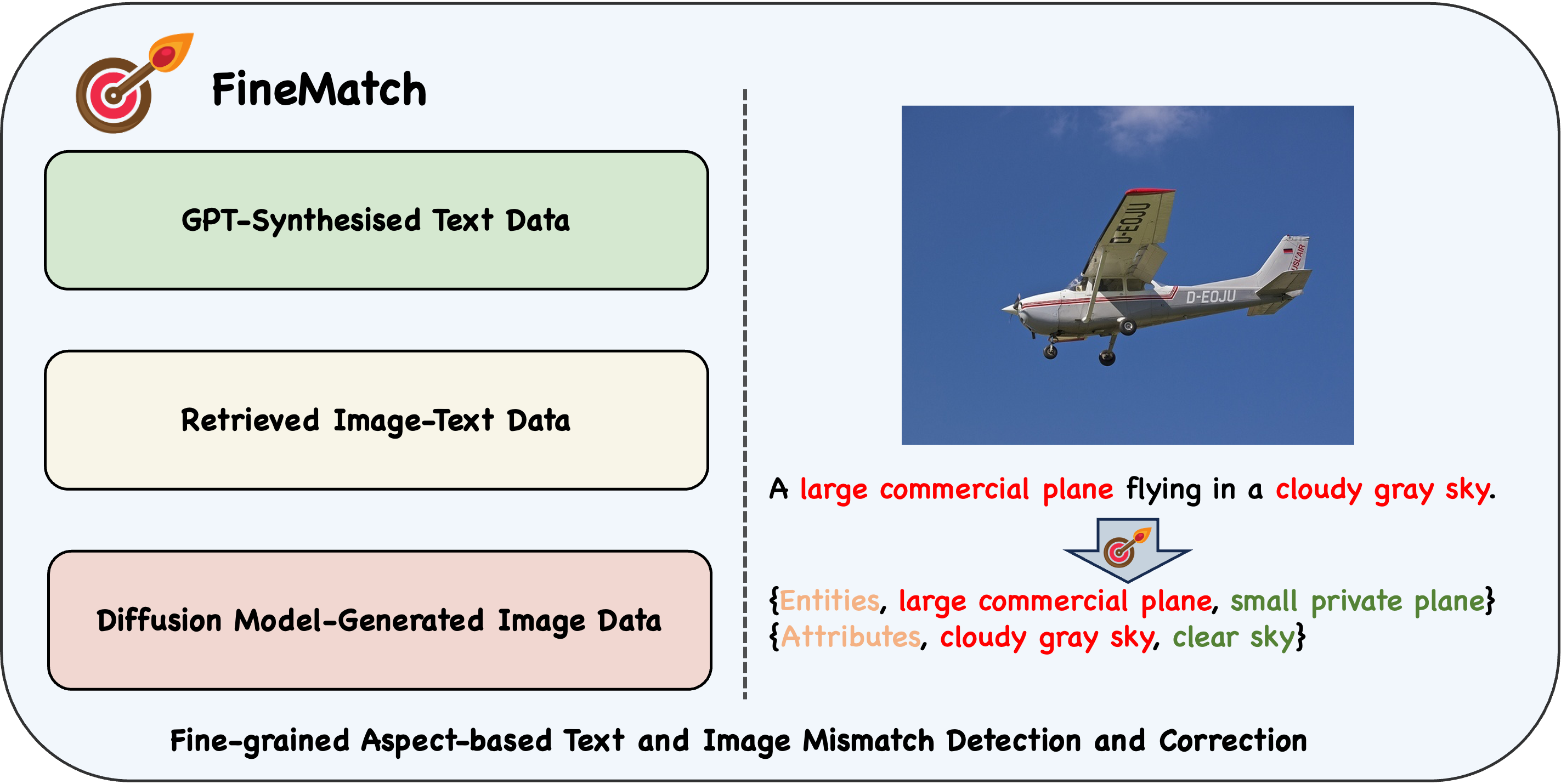}
  \caption{Given a text and image pair, \NAME enables VLMs to detect the \textcolor{textred}{\textbf{mismatched aspects}} and the \textcolor{textorange}{\textbf{aspect classes}} in the caption and then give the corresponding \textcolor{textgreen}{\textbf{corrections}}.}
  \label{fig:teaser}
  \vspace{-5mm}
\end{figure}

In recent years, there has been a growing focus on evaluating and improving the compositionality in large VL models \cite{diwan2022winoground,parcalabescu2021valse,parcalabescu2020seeing}. Most of these approaches focus on constructing hard negative text-image pairs to evaluate the models' compositionality \cite{zhao2022vl,thrush2022winoground,yarom2023you,he2023globalmapper,yuksekgonul2022and,hsieh2023sugarcrepe,ma2023crepe}. These evaluations typically require models to identify the hard negative samples at the sentence level but ignore evaluating the model's capability to localize the mismatched phrases and provide the corresponding corrections. Nevertheless, evaluating the compositionality of pretrained VL models from the perspective of sentence level may seem almost too trivial a task \cite{diwan2022winoground}. Additionally, research into the ability to identify and rectify discrepancies between images and text has been overlooked.
Based on this background, we propose \textbf{\NAME}, a new challenging benchmark for boosting VLMs' ability to identify and address the fine-grained discrepancies between visual and textual data.

With \NAME, we analyze and address the semantic discrepancies between visual and textual data from four aspects: \textbf{Entity}, \textbf{Relation}, \textbf{Attribute}, \textbf{Number}. And we provide examples for these four aspects in supplementary material.
We build \NAME data from both the image side and text side, aggregating data from multiple sources. The \NAME benchmark comprises 49,906 high-quality, human-annotated image-text pairs, distributed as 43,906 in the training set, 1,000 in the validation set, and 5,000 in the test set. This data originates from various sources, including: \textbf{GPT-Synthesized Text Data},
\textbf{Retrieved Image and Text Data}, and
\textbf{Diffusion Model-Generated Image Data}.
Each image-text pair encompasses a varying number of mismatched aspects, ranging from 0 to 3. The collection methods for each data source are described in Section \ref{sec:DC}.

To evaluate the ability to fine-grained text and image mismatch analysis of pretrained VLMs, we conduct experiments in both supervised learning and in-context learning settings. To verify the rationality of our proposed benchmark, we also provide the human performance results on the \NAME test set.

%\NAME benchmark contains 50k high quality human-annotated image and text pairs with 40k in training set, 2k in validation set, and 8k in test set. The data is collected from multiple source of data including: 1. GPT-synthesised query data; 2. Retrieved real-world data; 3. Diffusion model synthesised image data. Each text image pair contains different number of mismatch aspects from 0-3. \\

Our main contributions are three-fold:
\begin{itemize}
    \item We propose a novel task for aspect-based fine-grained image and text mismatch detection and correction. To support this endeavor, we have constructed a large-scale dataset \textbf{\NAME} with human annotations tailored to the proposed task. We put forth a new evaluation metric, \textbf{ITM-IoU}, which evaluates model predictions against ground truth on both the character and semantic levels. The experimental results show a high correlation between the ITM-IoU and human evaluation.
    
    \item We evaluate various state-of-the-art pre-trained VL models on our proposed benchmark. The empirical results show that training on \NAME can effectively improve the models' capability of identifying and rectifying text and image mismatches. 
    
    \item With \NAME, we are able to build a novel and simple self-correction text-to-image generation system, which can detect the detailed mismatch information between a generated image and text prompt and then automatically generate image editing instructions to edit the image to be semantically consistent with the text prompt. The generation examples indicate the system can effectively reduce hallucinations in text-to-image generation.

    %\item We provide a comprehensive evaluation of the ability of 
    
    %With FineMatch, we present a serious VL models that can perform  \textbf{Match-LLaVA}, the first model for aspect-based fine-grained text and image mismatch analysis based on visual instruction tuning. This model accepts an image and its corresponding caption as input and can predict mismatches between the text and the image. This model significantly improves the capability of identifying and rectifying text and image mismatches in Visual-Language (VL) models. We also demonstrate that this model can yield substantial benefits across a variety of downstream tasks. 
    
    %We provide a comprehensive evaluation of existing pre-trained VL models on our proposed benchmark. Our experiment results reveal a consistent correlation between the models' performance on downstream tasks and their ability to discern hard negative mismatch captions.
\end{itemize}
\section{Related Work}

\label{sec:relatedwork}
\subsection{Compositionality Evaluation}
Compositional image and text understanding is a critical capability for VLMs. Research indicates that VLMs struggle with distinguishing the hard negative examples, i.e., image text pairs that mismatch in at least one aspect (e.g., attribute, relation), since they have little incentive to
learn to encode compositionality during contrastive pretraining \cite{yuksekgonul2022and}. Moreover, finetuning with generated hard negative examples can improve the performance of language models \cite{hsieh2023sugarcrepe}. In recent years, numerous benchmarks have been proposed to assess the capability of VLMs for fine-grained compositional vision and language reasoning. VL-CheckList \cite{zhao2022vl} is an explainable framework that generates fine-grained and disentangled evaluation reports about VLMs. ARO \cite{yuksekgonul2022and} evaluate models for fine-grained relation, attribution, and order understanding.  Winoground \cite{thrush2022winoground} is a task for visio-logic compositional reasoning. SUGARCREPE \cite{hsieh2023sugarcrepe} aims to remove the artifact bias in model-synthesized visual compositional understanding evaluation benchmarks. Furthermore, there are several other benchmarks for compositionality evaluation, including SeeTRUE \cite{yarom2023you}, CREPE \cite{ma2023crepe}, Cola \cite{Ray2023COLAAB}, and T2I-CompBench \cite{Huang2023T2ICompBenchAC}, etc. However, there are no VL compositionality benchmarks for aspect-based fine-grained text and image match detection and correction. 
%is a framework to understand the capabilities to the compositional understanding of VLMs. The proposed method divides the image-texting ability of VLMs into three categories: objects, attributes, and relations, and uses a taxonomy to further break down these three aspects.
%A fundamental characteristic common to both human vision and natural language is their compositional nature.
%\subsection{Hallucination in VLMs}
 %Hallucination poses a significant challenge in VLMs, directly impacting the models' reliability. Prior works mainly focus on detecting and mitigating hallucinations  as well as evaluating hallucinations \cite{li2023evaluating,huang2023survey}.
 
%\cite{guan2023hallusionbench}

\subsection{Pretrained Vision-Language Models}
Large pretrained VL models such as OFA \cite{wang2022ofa}, BEiT-3 \cite{wang2022image}, CoCA \cite{yu2022coca}, and mPlug \cite{ye2023mplug} have successfully facilitated many cross-modal downstream tasks. CLIP \cite{radford2021learning} and its following works \cite{singh2022flava,li2022blip,mu2022slip} learn to align image and text features through contrastive learning objectives on large-scale image-text pairs. BLIP \cite{li2022blip,li2023blip}, LLaVA \cite{Liu2023VisualIT}, and MiniGPT4 \cite{chen2023minigptv2} show promising results by connecting vision encoders and LLMs through a compact intermediary model. These pretrained VL models with a stable fine-tuning strategy \cite{hua2022fine,hua2021noise} can be easily adapted to a new downstream task. In addition, GPT-4V \cite{OpenAI2023GPT4TR}, Flamingo \cite{alayrac2022flamingo}, and Emu2 \cite{sun2023generative} show strong abilities in zero-shot learning and multimodal in-context learning. Despite the remarkable achievements of large pretrained VL models, they struggle with capturing and understanding the fine-grained compositional information present in both text and images. The goal of this study is to provide a benchmark for evaluating and boosting the compositionality of pretrained VL models.

\section{Aspect-based Fine-grained Image and Text Mismatch Analysis}
\label{sec:DC}
\subsection{Task Definition}
\NAME contains two subtasks: (1) \textbf{Mismatch Detection  (MD)}; (2) \textbf{Mismatch Detection \& Correction (MD\&C)}. Let $\mathcal{D}=\{I_i,C_i,P_i\}_{i=1}^{|\mathcal{D}|}$ represents the dataset. $(I_i,C_i)$ is the image and text pair, and $P_i=\{c_j,p_j,o_j\}_{j=1}^M$ is the mismatched aspects representation, where $c_j$ is the aspect class, $p_j$ is the aspect phrase that is extracted from the caption $C_i$, $o_j$ is the corresponding correction, $M$ is the number of mismatched aspect. Given an image-text pair $(I_i,C_i)$, for the Mismatch Detection task, the models need to predict a set of tuples that contains the mismatched phrases in the captions and the corresponding class.
\begin{dmath}
\textbf{MD}(I_i,C_i)
=\{(c_j,p_j)\}_{j=1}^M
\end{dmath}
For the Mismatch Detection \& Correction task, the models need to predict a set of triplets that contains the mismatched phrases, the class of the phrase, and the suggested corrections.
\begin{dmath}
\textbf{MD\&C}(I_i,C_i)
=\{(c_j,p_j,o_j)\}_{j=1}^M
\end{dmath}

\NAME also contains the data for which the caption matches the image. In this case, the models' output should be $None$.

\subsection{GPT-Synthesized Text Data with Aspect Graph Parsing and Node Replacement}

\subsubsection{Data Generation}
\begin{figure}[t!]
  \centering
  \includegraphics[width=0.9\linewidth]{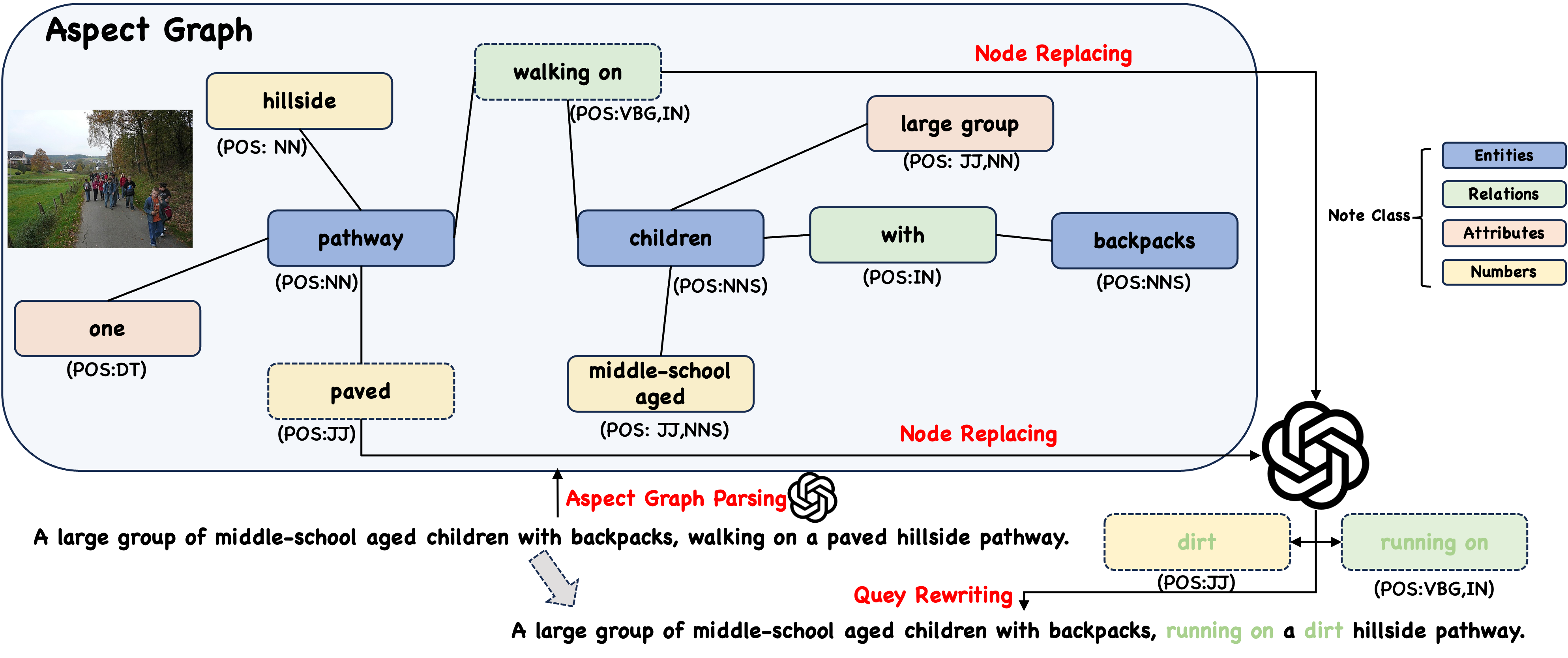}
  \caption{Aspect graph parsing and node replacement for GPT-Synthesized text data.}
  \label{fig:graph}
  \vspace{-0.4cm}
\end{figure}
For the GPT-Synthesised text data, we generate mismatched captions via aspect graph parsing and node replacement. To extract fine-grained compositional sub-phrases from the captions, we parse the captions into aspect graphs using In-Context Learning (ICL) with GPT-4. As depicted in Figure \ref{fig:graph}, an aspect graph consists of nodes representing aspect entities and edges illustrating the relationships between these entities, with each node being atomic. Then, we prompt GPT-4 to randomly replace the nodes with the counterfactual descriptions while maintaining the same Part of Speech (POS) tag with the initial nodes. Subsequently, the aspect graphs are translated back into new captions. This approach enables us to create mismatched captions without changing the structure of the initial captions. We filter the generated captions with the CLIP score, to remove any mismatched captions that are significantly incongruent with the images. 
%\vspace{-2mm}
\subsubsection{Data Debiasing}
Previous work \cite{hsieh2023sugarcrepe} pointed out that the rule-based mismatched caption generation procedures may introduce two major types of undesirable artifacts: (1) nonsensible artifacts (irrational contents), and (2) non-fluent artifacts (grammar issues). Additionally, the significance of the semantic gap between mismatched captions and their associated images is another artifact bias for the generated data, making these captions easy to discern by models. To fix these biases and ensure the quality of the GPT-Synthesised text data, we first use the combined Vera score \cite{liu2023vera}, grammar score \cite{morris2020textattack}, and CLIP score \cite{radford2021learning} to filter the data. This approach allows us to exclude examples that exhibit grammatical errors, contradict common sense, or present significant discrepancies between the image and caption. Subsequently, the filtered data is annotated by human experts. In the annotation, the workers are required to check and revise the GPT-synthesised captions and the corresponding mismatched aspects. The quality analysis indicates this aspect-graph parsing and node-replacing method combined with human annotation can effectively reduce the artifact biases in the GPT-synthesised data.

\subsection{Retrieved Image-Text Data}
The semantic discrepancy between text queries and images is a common issue in text-to-image retrieval systems. We can utilize this property of text-to-image retrieval systems to obtain the mismatched text and image pairs. We select text queries with rich compositional structures from various datasets, including NoCaps \cite{Agrawal2019nocapsNO} and WizViz \cite{gurari2020captioning}. To get the text queries with rich compositional information, we first parse each text query into a constituency tree using syntactic parsing tools SpaCy and then filter the queries according to the depth of the tree. Queries with a deeper constituency tree indicate the more complex syntax of the sentence, and these sentences contain more compositional information. We sample 10k diverse and complex queries for image retrieval. The retrieved images source include: LAION-400M \cite{Schuhmann2022LAION5BAO}, COYO-700M \cite{kakaobrain2022coyo-700m}, and Smithsonian Open Access \cite{smithsonian_open_access_2023}. We first use the ViT-G/14 CLIP model with the weight from Open CLIP \cite{ilharco_gabriel_2021_5143773} to retrieve 10 candidates from the image datasets and then filter the images with the combination of aesthetic score, similarity score, and image size. We finally obtain 10K high-quality text and image pairs for human annotation.

\subsection{Stable Diffusion Generated Image Data}
The text query comes from T2I-CompBench \cite{Huang2023T2ICompBenchAC}, a benchmark for compositional text-to-image generation. All the text queries in the benchmark are meticulously designed in 3 categories (attribute
binding, object relationships, and complex compositions). We use the text queries from the T2I-CompBench training set to prompt Stable Diffusion 2.1 \cite{Rombach_2022_CVPR} to generate images. We finally obtain 2.5K high-quality text and image pairs for human annotation.

%\subsection{Data Filtering}
%Prior work revels \cite{hsieh2023sugarcrepe} that the GPT-synthesised data introduces undesirable artifact bias which could result in nonsensical and non-fluent in grammar for the generated mismatched text. These bias renders the dataset hackable. To fix the bias, we filter the GPT-synthesised data using the Vera score \cite{liu2023vera}  and grammar score \cite{morris2020textattack} to filter out the data points with significant nonsensical plausibility and low grammar scores. 

\subsection{Human Annotation}
To standardize the labeling scheme across different data sources and ensure data quality while eliminating potentially harmful content and addressing ethical concerns, we employ a consistent annotation team composed of the same group of workers. The annotation interfaces of different sources of data are shown in Appendix Section 7.

\subsection{Comparison of \textbf{\NAME} with Previous Works}

We summarize the novelty of our work compared with previous works.
Our work introduces a novel task centered on aspect-based, fine-grained detection and correction of text and image mismatches. Previous research mainly focuses on evaluating the pre-trained models’ ability to identify the hard negative examples \cite{hsieh2023sugarcrepe,ma2023crepe,yuksekgonul2022and} via retrieval accuracy and ignores evaluating models' ability to detect which part of the text is mismatched with the image and correct the mismatched aspects. 
Moreover, earlier studies, such as ARO \cite{yuksekgonul2022and}, predefined the instance classes and collected images with 48 relations and 117 attribute pairs, thereby constraining the diversity of instances. In contrast, our proposed benchmark achieves the open set in text and image mismatch detection. Furthermore, previous works that perform fine-grained text and image mismatch detections depend on the VQA-based method, which needs to generate a list of questions first and then employ the VQA model to answer the questions. These methods, however, are not flexible and may suffer from bias accumulation issues. Our method trains models to generate the mismatched aspect phrase and the correct aspect phrase in an end-to-end manner. In Table \ref{table:dataset_comparison}, we also compare the difference of \NAME with other related works from the perspective of the size of the dataset, whether fine-grained IMT detection and correction is supported, and if human annotation is employed to improve the data quality and remove harmful content.

\begin{table*}[t!]
\centering
\caption{Comparison of Different Datasets}
\label{table:dataset_comparison}
\resizebox{\linewidth}{!}{
\begin{tabular}{@{}l|c|c|c|c@{}} % Alignment: left, center, center, center
\toprule
\textbf{Benchmark} & \textbf{\# Images-Text Pair} & \textbf{Fine-grained Mismatch Detection \& Correction}& \textbf{Human Annotation} & \textbf{Multiple Source/Domain}  \\
\midrule

ARO \cite{yuksekgonul2022and} & 28,748 & \redcheck  &\redcheck  & \redcheck\\
SUGARCREPE \cite{hsieh2023sugarcrepe} & 7512 & \redcheck  &\greencheck &\redcheck \\
Winoground \cite{thrush2022winoground}& 800 & \redcheck  &\greencheck &\redcheck \\
CREPE \cite{ma2023crepe} & 370,000 & \redcheck  &\redcheck &\greencheck\\
VL-Checklist \cite{zhao2022vl} & 410,000 & \redcheck  &\redcheck &\greencheck \\
SeeTrue \cite{yarom2023you} & 31,855 & \redcheck  &\greencheck&\greencheck \\
\midrule
\textbf{\NAME} & 49,906 & \greencheck  & \greencheck &\greencheck \\
% Add more rows here as needed
\bottomrule
\end{tabular}}
\vspace{-4mm}
\end{table*}

%\vspace{-1mm}
\subsection{Evaluation}
\label{sec:eval}
To evaluate models' performance on \NAME, we propose a novel metric called ITM-IoU. IoU (Intersection over Union) is a standard metric in computer vision for measuring the accuracy of object detection or segmentation, based on the overlap between predicted and ground truth boundaries or pixels. In this study, since each caption may contain multiple mismatched aspects, we compute the IoU between the models' predicted set of aspect triplets and the set of ground truth triplets. For triplets matching, we draw inspiration from the generic structured prediction evaluation methods in the NLP field \cite{Lu2024WebLINXRW}. To evaluate the accuracy of the predicted mismatched aspect classes, we adopt the exact match (EM) \cite{rajpurkar2016squad} metrics. For mismatched aspect phrase detection, we evaluate from both the character level and semantic level. For the lexical similarity evaluation, we use chrF \cite{Popovic2015chrFCN}, an F1-score for character n-gram matches (we use the default setting of n = 6). For the semantic level evaluation, we use the BERT score \cite{zhang2019bertscore}. Given a predicted mismatched aspect representation $P_i=\{c_j,p_j,o_j\}_{j=1}^M$ ($i \in \{1, 2, ..., |\mathcal{D}|\}$) and the corresponding ground truth $G_i=\{c_j^\prime,p_j^\prime,o_j^\prime\}_{j=1}^{M^\prime}$ ($M^\prime$ is the number of ground truth mismatched aspects), the combined detection score ${Score_D}_j$ is calculated as:
\begin{equation}
    \mathit{Score_{\mathit{D}}}_j=\frac{\mathit{BERTScore(p_j, p_j^\prime)}+chrF(p_j, p_j^\prime)}{2},
\end{equation}
As the aspect phrase correction is an open-ended generation task, we calculate the BERT score to evaluate the semantic similarity of the generated corrections $o_j$ and the ground truth $o_j^\prime$. The correction score ${Score_{\mathit{C}}}_j$ is calculated as:
\begin{equation}
    \mathit{Score_{\mathit{C}}}_j=\mathit{BERTScore(o_j, o_j^{\prime})},
\end{equation}
The total score of a predicted aspect is the weighted sum of the three elements' scores in the mismatched aspect representation:
\begin{equation}
\begin{aligned}
\mathit{Score_{\mathit{Aspect}}}_j=&W_{\mathit{Ca}}\cdot EM(c_j,c_j^\prime)+&W_{\mathit{De}} \cdot \mathit{Score_{\mathit{D}}}_j+ W_{\mathit{Co}} \cdot \mathit{Score_{\mathit{C}}}_j
\end{aligned}
\end{equation}
where the $W_{\mathit{Ca}},\ W_{\mathit{De}}$, and $\ W_{\mathit{Co}}$ is the weight of the $EM, \ \mathit{Score_{\mathit{D}}}_j$, and $ \mathit{Score_{\mathit{C}}}_j$, respectively.
In this study, we set the weights $W_{\mathit{Ca}}=0.2$, $W_{\mathit{De}}=0.4$, and $W_{\mathit{Co}}=0.4$. To compute the IoU of the aspect representation, we set a threshold $T$ to match the predictions with ground truth, if $\max(\{\mathit{Score_{\mathit{Aspect}}}_k\}_{k=1}^{M^\prime}) \geq T$ then the predicted triplet matches the ground truth.  For each aspect representation prediction, we compute the final score as:
\begin{equation}
    \mathit{Score_{\mathit{Aspect}}}_j = \left\{\begin{array}{lcl}
        \mathit{Score_{\mathit{Aspect}}}_k & \text{if} & \max(\{\mathit{Score_{\mathit{Aspect}}}_k\}_{k=1}^{M^\prime}) \geq T, \\
        0 & \text{else} & 
    \end{array}\right.
\end{equation}
Then the final calculation of the mismatched aspect ITM-IoU is calculated as:
\begin{equation}
   \text{ITM-IoU} =\frac{\sum_{j=0}^{M} \mathit{Score_{\mathit{Aspect}}}_j}{M} \times \frac{|P_i \cap G_i|}{|P_i \cup G_i|},
\end{equation}
where $|P_i \cap G_i|$ denotes the number of matched triplets of data $i$, and $|P_i \cup G_i| = |P_i| + |G_i| - |P_i \cap G_i|$. We also provide the pseudocode in supplementary material for the ITM-IoU calculation. 

A common way to show the goodness of an evaluation metric is to show its correlation with human evaluations. We conduct a human evaluation for the experimental results in Section \ref{sec:human}, the results indicate the high correlation of ITM-IoU with human evaluation, which reflects the rationality of our proposed evaluation metrics.
\begin{figure}[htbp!]
  \centering
  \includegraphics[width=0.49\linewidth]{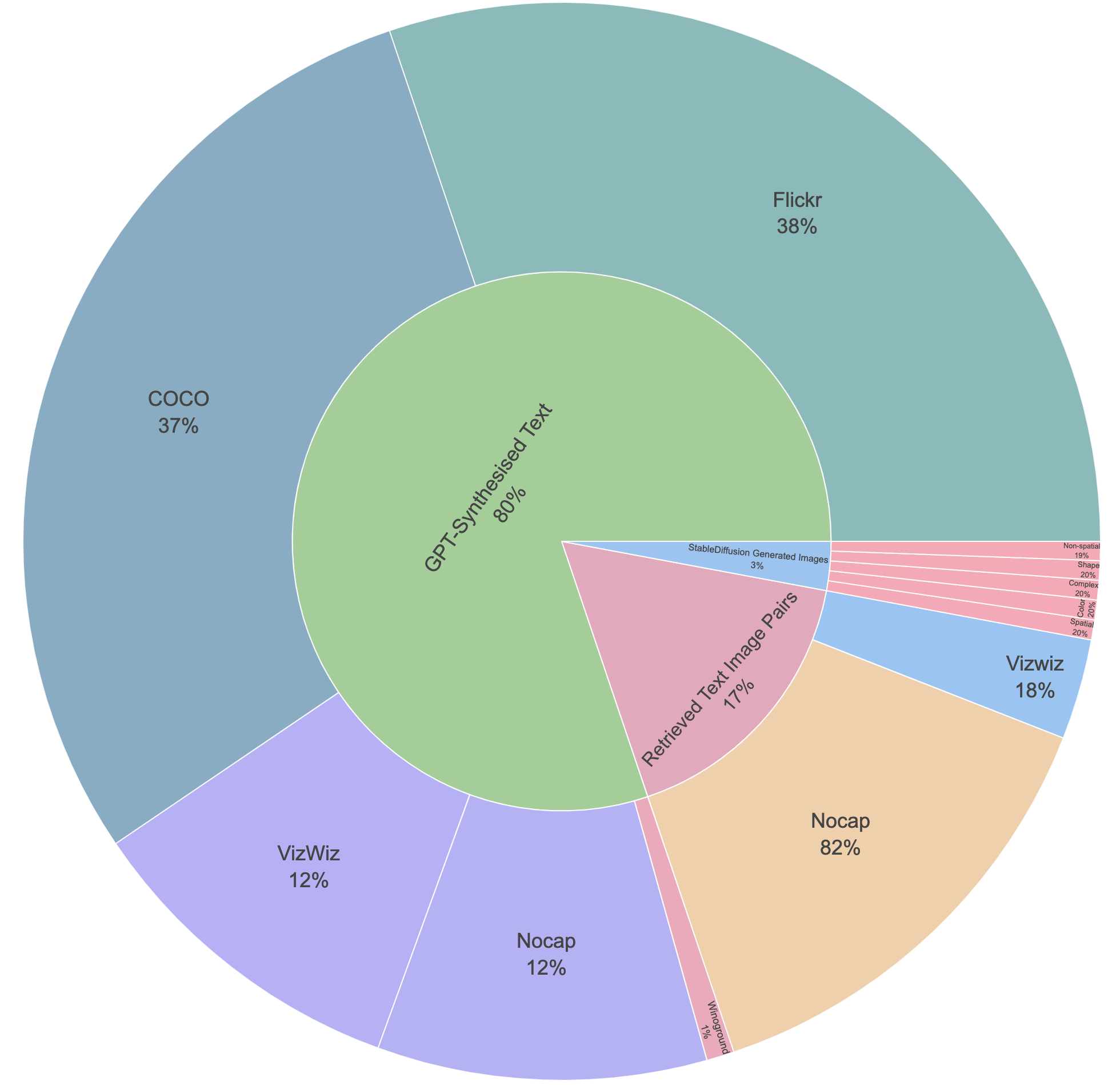}
  \includegraphics[width=0.5\linewidth]{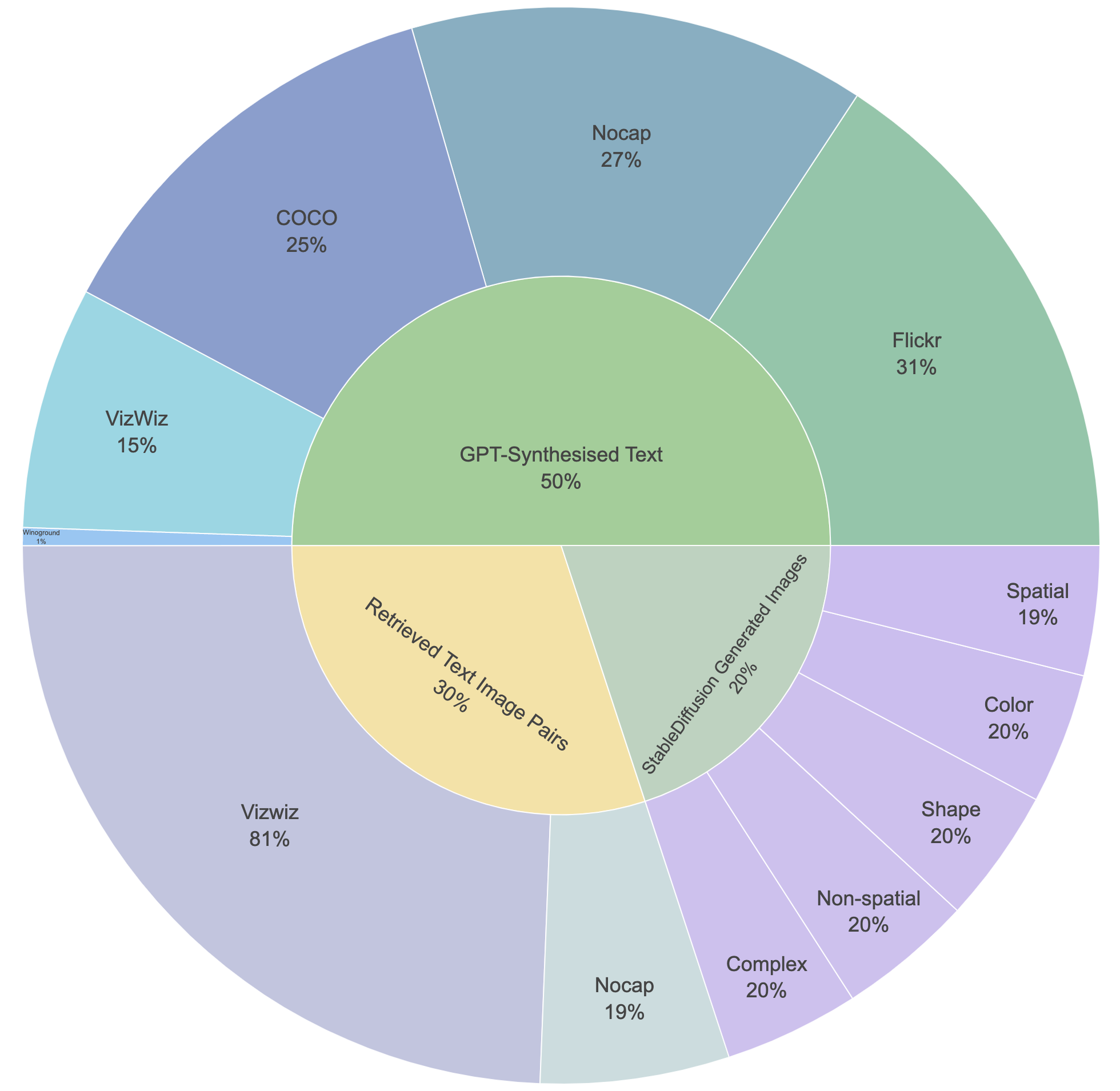}
  \caption{The initial data source distribution (inner circle) and domain distribution (outer circle) for the \NAME training set (left) and test set (right).}
  \label{fig:sunburst}
\end{figure}

\subsection{Quantitative Analysis}

%In this section, we present quantitative analysis for the \NAME dataset. 

\subsubsection{Distribution of Data Source and Domain}
We provide the text and image distribution analysis from different perspectives. 
Figure \ref{fig:sunburst} shows the data source distribution in the inner circle and domain distribution in the otter circle of \NAME training and test set.

\begin{figure*}[htbp]
  \centering
  \includegraphics[width=\textwidth]{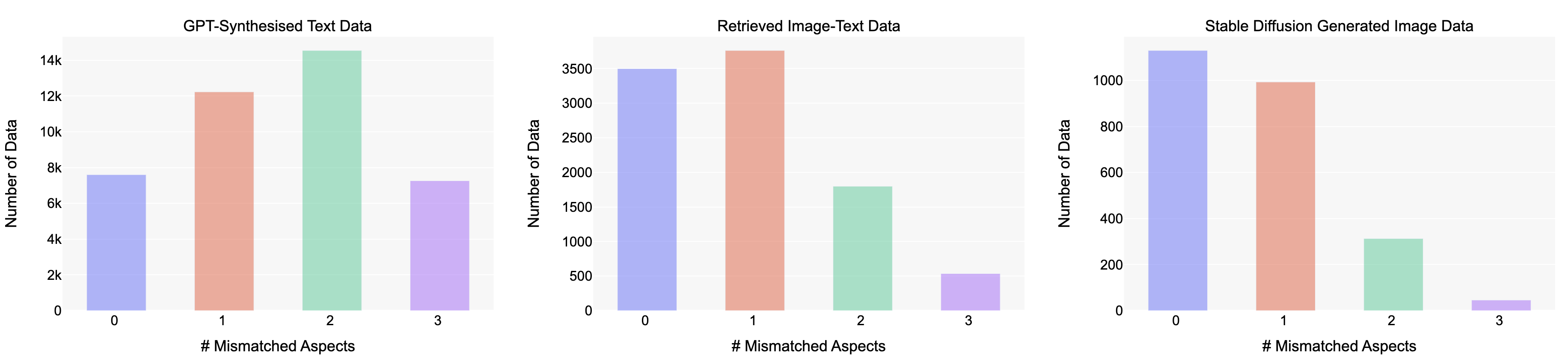}
  \caption{Data distribution of varying numbers of mismatched aspects across different data sources in \NAME.}
  \label{fig:histogram}
  \vspace{-2mm}
\end{figure*}

\subsubsection{Distribution of Mismatched Aspects}
Figure \ref{fig:histogram} shows the number of mismatched aspects distribution across different data sources in \NAME.

\begin{figure*}[htbp]
  \centering
  \includegraphics[width=\textwidth]{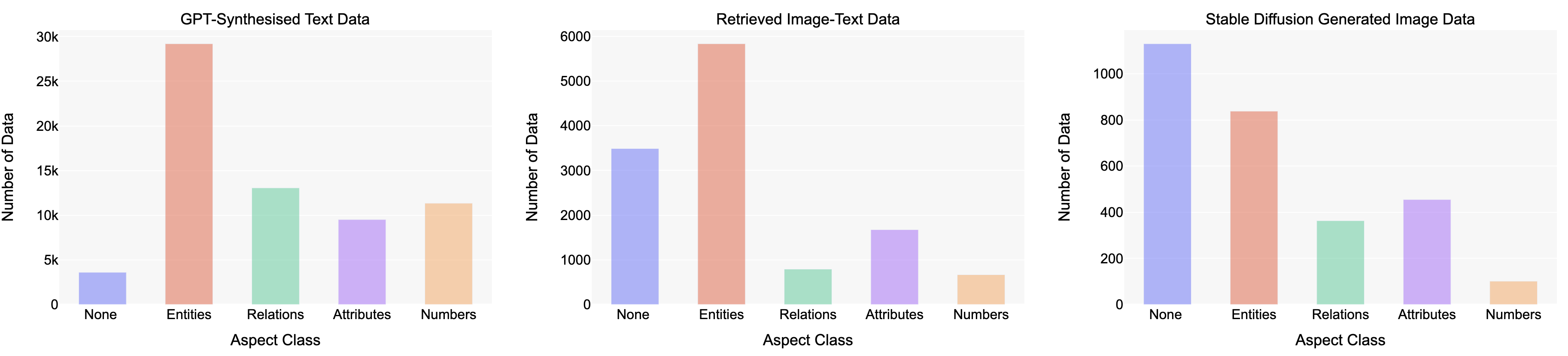}
  \caption{Data distribution of the mismatched aspect classes across the training, validation, and test sets in \NAME.}
  \label{fig:class}
\end{figure*}
  \vspace{-0.2cm}
\subsubsection{Distribution of Aspects Classes}
Figure \ref{fig:class} shows the distribution of aspect classes across different data sources in \NAME.

\subsection{Qualitative Analysis}
\begin{figure*}[htbp]
  \centering
  \includegraphics[width=\textwidth]{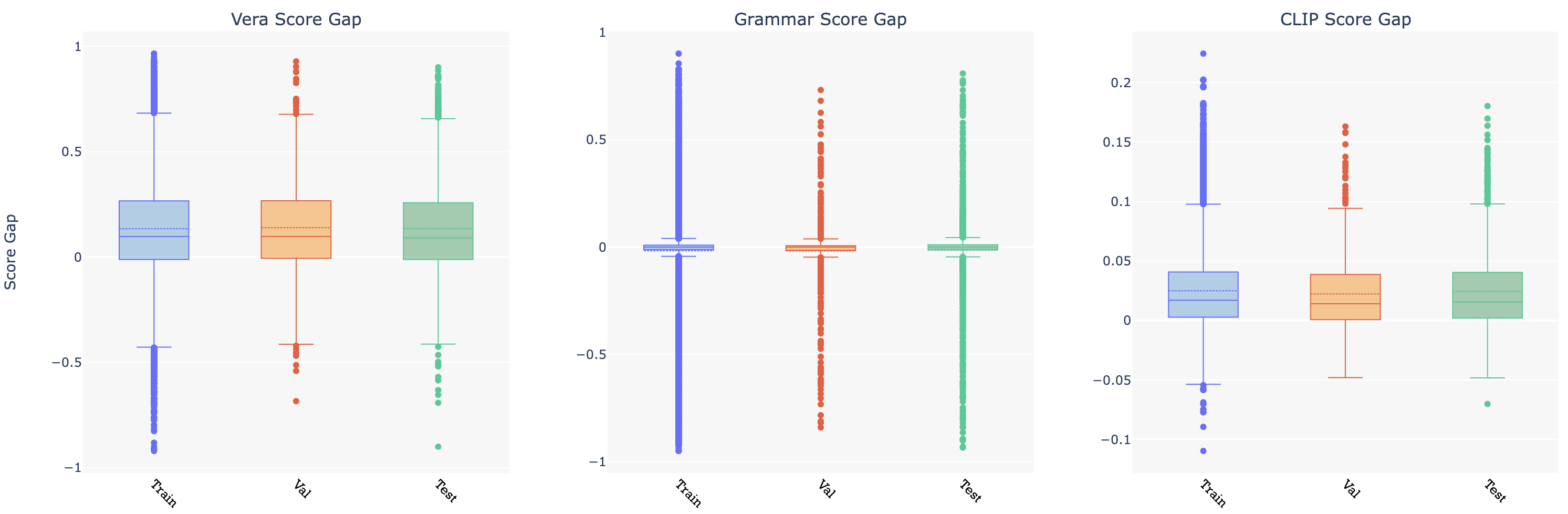}
  \caption{Score gap distribution in \NAME. Since we use human annotation to replace the adversarial refinement, the score gap distribution lies on the positive spectrum, but these effects are within an acceptable range.}
  \label{fig:score}
  \vspace{-5mm}
\end{figure*}

We present the analysis of human-annotated and GPT-Synthesized data, focusing on the Vera Score Gap, Grammar Score Gap, and CLIP Score Gap. These metrics compare the quality changes of human-annotated mismatched captions against the original captions. Previous research \cite{hsieh2023sugarcrepe} analyzes the artifact bias through the Vera Score Gap and Grammar Score Gap, and employs the adversarial refinement to fix this bias. In this research, we utilize human annotation to fix this bias and add the CLIP Score Gap to quantify the semantic changes of the human-annotated mismatched captions. The score gap is typically calculated as $S^P-S^N$, where the $S^P$ and $S^N$ are the Vera/Grammar/CLIP scores of the initial captions and the human-annotated GPT-Synthesised data. The findings, illustrated in Figure \ref{fig:score}, reveal that score gap distribution lies on the positive
spectrum, indicating that hard negative samples in the GPT-Synthesized data exhibit a higher likelihood of being nonsensical. The results also reflect a marginally reduced similarity to the initial matching captions. While these effects are within an acceptable range. Moreover, GPT rewrites do not significantly impact grammar fluency or introduce errors compared to the original captions. We also provide more quality analysis from different perspectives about other data sources in Appendix Section 2.

\vspace{-5mm}

% We leverage the grammar-check model \cite{morris2020textattack} that assigns high scores to grammatically correct texts to capture the non fluent bias of human annotated data. We then utilize Vera \cite{liu2023vera}, a plausibility estimation model, to characterize the nonsensical bias. The analysis results are shown in Figure \ref{}.

%We filter out the annotations that gain low Vera score and grammar-check score. 
%The results show that through human annotation, the artifact bias of the annotated data could be effectively reduced. 

\section{Experiments}
\subsection{Visual Instruction Tuning}

We train the models on \NAME in an image-to-text generation setting. Given an image $I_i$ and the corresponding caption $C_i$, and the output aspect representations $P_i=\{c_j,p_j,o_j\}_{j=1}^M$, the training objective is:
\begin{equation*}
\mathcal{L} =- \sum_{\mathcal{D}} \sum_{t=1}^M \log p (P_t\mid [C_i:I_i], P_{\leq t-1}).
\end{equation*}

We conduct experiments on various state-of-the-art pretrained VL models including OFA \cite{wang2022ofa}, \cite{ye2023mplug}, InternLM-Xcomposer2-VL \cite{internlmxcomposer2}  LLaMA-Adapter2 \cite{gao2023llamaadapterv2}, MiniGPT-4 \cite{chen2023minigptv2}, ShareGPT4V \cite{chen2023sharegpt4v}, and LLaVA series \cite{Liu2023VisualIT}. The experiment results are shown in Table \ref{tab:mainresults}.
\vspace{-4mm}
\begin{table*}[htbp]
\small
\centering
\caption{
Visual instruction tuning performance (ITM-IoU) of different VL models on the \NAME test set.
}
% \begin{tabular}{P{1.7cm}P{1.2cm}P{1.2cm}P{1.0cm}P{1.2cm}P{1.2cm}P{1.2cm}P{1.1cm}P{1.1cm}P{1.1cm}}
\resizebox{\linewidth}{!}{
\begin{tabular}{l|c|c|c}
\toprule[1.2pt]
\textbf{Models}&\textbf{Size} & \textbf{Mismatch Detection $\uparrow$} & \textbf{Mismatch Detection\&Correction $\uparrow$}\\
\midrule
OFA-Large \cite{wang2022ofa}&472M&19.72 &21.35 \\
LLaMA-Adapter2 \cite{gao2023llamaadapterv2}&7B &35.84&40.76\\
mPLUG-Owl2 \cite{ye2023mplug} &8.2B&46.70  &48.28 \\
MiniGPT-4-V2 \cite{chen2023minigptv2}  &7B&51.18 &55.95 \\
InternLM-Xcomposer2-VL \cite{internlmxcomposer2} &7B&58.70&61.07\\
LLaVA-1.5-LoRA \cite{Liu2023VisualIT}&7B&62.18  &63.80\\
LLaVA-1.5 \cite{Liu2023VisualIT}&7B&62.25  &63.62\\
LLaVA-1.5-LoRA \cite{Liu2023VisualIT}&13B &65.51  & 66.73\\
LLaVA-1.5 \cite{Liu2023VisualIT}&13B &66.02  &67.13\\
ShareGPT4V \cite{chen2023sharegpt4v} &13B &66.06 &67.21 \\
LLaVA-1.6-Vicuna \cite{liu2024llavanext} &13B &66.10  &67.31\\
\midrule
\rowcolor{mygray}
\textbf{Human Performance} &-  &\bf{88.32}  & \bf{89.19}\\
\bottomrule[1.2pt]
\end{tabular}
}
\label{tab:mainresults}
% \bottomrule
 \vspace{-5mm}
\end{table*}
%From Table \ref{tab:mainresults}, we can summarize that the VL models with larger language models (e.g., LLaVA-1.5 7B and LLaVA-1.5 13B) performs better on \NAME. The language models pretrained on more training data or carefully designed instruction following data (e.g., ShareGPT4V V.S. LLaVA-1.5 ) achieves better performance. In addition, models improved with reasoning, image encoder, and world knowledge (LLaVA-1.6 and LLaVA-1.5) can obtain more performance gains.

From the results presented in Table \ref{tab:mainresults}, it is evident that VLMs integrated with larger language models (LLaVA-1.5 7B vs. LLaVA-1.5 13B) exhibit superior performance on \NAME. Language models that have been pretrained on more data or those that have been finetuned with carefully designed data  (e.g., ShareGPT4V vs. LLaVA-1.5) tend to achieve enhanced performances. Furthermore, models with improved reasoning capabilities, image encoder, and world knowledge ( LLaVA-1.6 vs. LLaVA-1.5) also demonstrate performance improvements. 

\subsection{In-Context Learning}
In-Context Learning (ICL) explores training-free
few-shot learning, where models are encouraged to “learn to
learn" from limited tasks and generalize to unseen tasks. In this study, we conduct experiments on public accessible pretrained VL models including MMICL \cite{zhao2023mmicl}, Otter \cite{li2023otter}, OpenFlamingo \cite{awadalla2023openflamingo}, Emu2 \cite{sun2023generative}, Gemini Pro Vision \cite{team2023gemini}, and GPT4-V \cite{OpenAI2023GPT4TR}. The experiment results are shown in Table \ref{tab:zsl}. Since GPT-4V and Gemini cannot process some specific contents in the image of \NAME test set, which will be judged as illegal input, we discard these examples, and we obtain 4278/5000 for GPT-4V and 4824/5000 for Gemini Pro to calculate the ITM-IoU.

\begin{table*}[htbp]
\small
\centering
\caption{
In-context learning results (in terms of ITM-IoU) of different VL models on the \NAME test set. ($\text{}^*$ indicates subset of the \NAME test set).
}
% \begin{tabular}{P{1.7cm}P{1.2cm}P{1.2cm}P{1.0cm}P{1.2cm}P{1.2cm}P{1.2cm}P{1.1cm}P{1.1cm}P{1.1cm}}
\resizebox{\linewidth}{!}{
\begin{tabular}{l|c|c|c}
\toprule[1.2pt]
\textbf{Models}&\textbf{Size} & \textbf{Mismatch Detection $\uparrow$} & \textbf{Mismatch Detection\&Correction $\uparrow$}\\
\midrule

Otter \cite{li2023otter}&7B&0.03&0.09\\
MMICL \cite{zhao2023mmicl}&7B&0.11&0.25\\
OpenFlamingo \cite{awadalla2023openflamingo}&9B&0.34&0.96\\
Emu2 \cite{sun2023generative}&37B&6.10&11.23\\
\rowcolor{mygray}
$\text{Gemini Pro Vision}^*$ \cite{team2023gemini}&- &9.07 &11.14 \\
%\midrule
\rowcolor{mygray}
$\text{GPT-4V}^*$ \cite{OpenAI2023GPT4TR}&- &\bf{21.92} &\bf{21.58} \\

\bottomrule[1.2pt]
\end{tabular}
}
\label{tab:zsl}
% \bottomrule
 \vspace{-5mm}
\end{table*}
It can be summarized from Table \ref{tab:zsl} that even the models such as GPT-4V and Gemini Pro Vision, which have a strong ability of multimodal in-context learning, are not as skilled at fine-grained compositional image and text matching analysis as we might have expected. Compared with the results in Table \ref{tab:mainresults}, models trained on \NAME demonstrate enhanced proficiency in detecting fine-grained text and image mismatches.

\subsection{Human Evaluation}
\label{sec:human}
To evaluate the rationality of our proposed ITM-IoU, we carried out a human evaluation of model predictions on the sampled \NAME test set. The sample size is 500. The human annotators are required to rate the quality of the generated mismatched aspect representations, the score ranges from 1-5, and the higher the better. The findings, presented in Table \ref{tab:HM}, demonstrate a high correlation between human evaluation results and the automatic evaluation metric ITM-IoU detailed in Table \ref{tab:mainresults}. 
\vspace{-4mm}
\begin{table}[htbp]
\small
\centering
\caption{
Average human evaluation scores (ranging form 1-5) for fully supervised-learning (the upper row) and in-context learning methods (the lower row).
}
% \begin{tabular}{P{1.7cm}P{1.2cm}P{1.2cm}P{1.0cm}P{1.2cm}P{1.2cm}P{1.2cm}P{1.1cm}P{1.1cm}P{1.1cm}}
\resizebox{0.8\linewidth}{!}{
\begin{tabular}{l|c|c}
\toprule[1.2pt]
\textbf{Model} & \textbf{Mismatch Detection $\uparrow$} & \textbf{Mismatch Detection\&Correction $\uparrow$}\\
\midrule
mPLUG-Owl2 &3.56  & 3.41 \\
MiniGPT-4 &3.77  &  3.65 \\
LLaVA-1.5-7B &4.02  & 3.89  \\
LLaVA-1.5-13B &4.41  &4.15 \\
\midrule
OpenFlamingo &2.01 &1.63 \\
Emu2 &2.55 &2.32 \\
Gemini Pro Vision&2.95 & 2.86\\
GPT4-V &3.35 & 3.27 \\
\bottomrule[1.2pt]
\end{tabular}
}
\label{tab:HM}
% \bottomrule
 \vspace{-5mm}
\end{table}
\vspace{-4mm}
\subsection{Failure Cases of Language Models on \NAME}
\begin{figure}[htbp]
  \centering
  \includegraphics[width=0.9\linewidth]{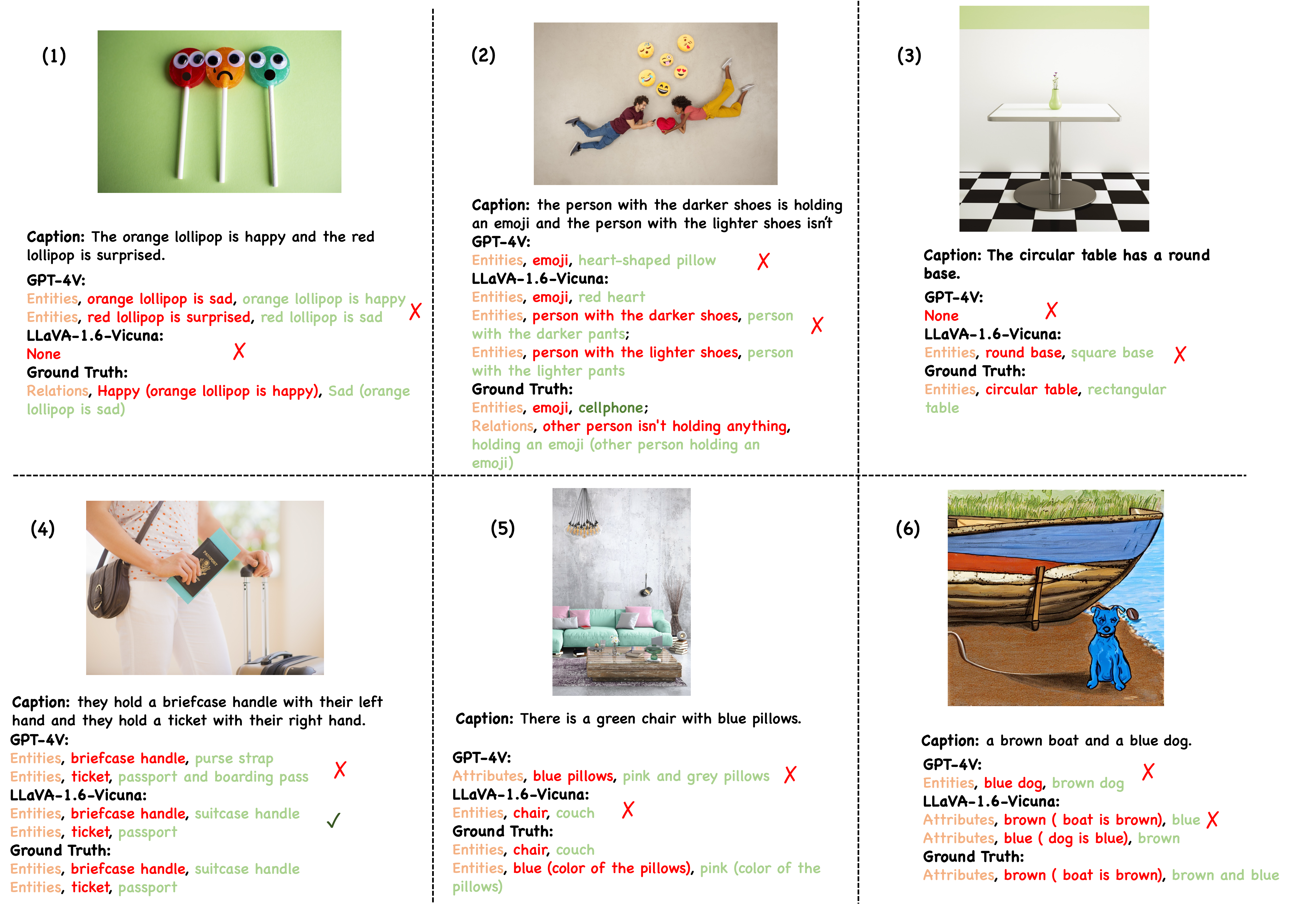}
  \caption{Failure cases (\{\textcolor{textorange}{\textbf{aspect classes}},\textcolor{textred}{\textbf{mismatched aspects}},\textcolor{textgreen}{\textbf{corrections}}\}) of GPT-4V and LLaVA-1.6 on \NAME.}
  \label{fig:fcase}
   \vspace{-5mm}
\end{figure}

In Table \ref{tab:mainresults} and Table \ref{tab:zsl}, we show the performance (in terms of ITM-IoU) of current state-of-the-art language models on \NAME. The results indicate that the models still have difficulties in identifying the mismatched aspects of the given data. We also show examples of the GPT4-V's predictions, the finetuned LLaVA-1.6 predictions, and the ground truth in Figure \ref{fig:fcase} to illustrate that fine-grained image-text matching is a challenging task for VLMs. 

\section{\NAME for Text-to-Image Generation Hallucination Detection and Correction}
% \subsection{Auto Correction of Text-to-Image Generation}
% Through \NAME, we can train models to find and correct the mismatch between text and image. We demonstrate how to use \NAME finetuned model to detect the mismatch of text and image and generate image editing prompts. Specifically, we use the finetuned LLaVA-v1.5-13B as the mismatche detection and correction module, and then use the GPT-4 as the image editing prompt generation module, finally, a InstructPix2Pix \cite{brooks2023instructpix2pix} model is employed to perform image editing.
Hallucination issues are prevalent in text-to-image (T2I) generation models, particularly for the text prompts that describe detailed scenes and intricate relationships~\cite{rawte2023survey, liu2024survey}. Many approaches are proposed for reducing hallucination for T2I models \cite{liu2023aligning,wang2023vigc,chen2024textdiffuser,song2023emotional,song2022objectstitch,yin2023woodpecker}.
In this study, we illustrate how \NAME can help detect and address the mismatch between the text prompts and the generated images for T2I generation models.

To achieve this goal, we develop a framework named \AT that incorporates \NAME with T2I models and VLMs for T2I automatic hallucination detection and correction. This system comprises four key modules: the T2I generation module, the hallucination detection module, the image editing prompt generation module, and the image editing module. Given a text prompt $T$ and its corresponding generated image $I$, the text-image pair is evaluated by a VLM (LLaVA-1.6 Vicuna) fine-tuned with \NAME to identify if there are any discrepancies between $I$ and $T$. Upon detecting a mismatch, the system invokes the image editing prompt generation module (utilizing GPT-4 for this purpose) to generate an image editing prompt. Subsequently, the image editing module (employing MagicBrush~\cite{zhang2024magicbrush} for image editing) is engaged to adjust the image. This process of hallucination detection, generation of editing prompts, and image editing is iteratively performed until the edited image satisfactorily aligns with the text prompt. 

\begin{figure}[htbp]
  \centering
  \includegraphics[width=\linewidth]{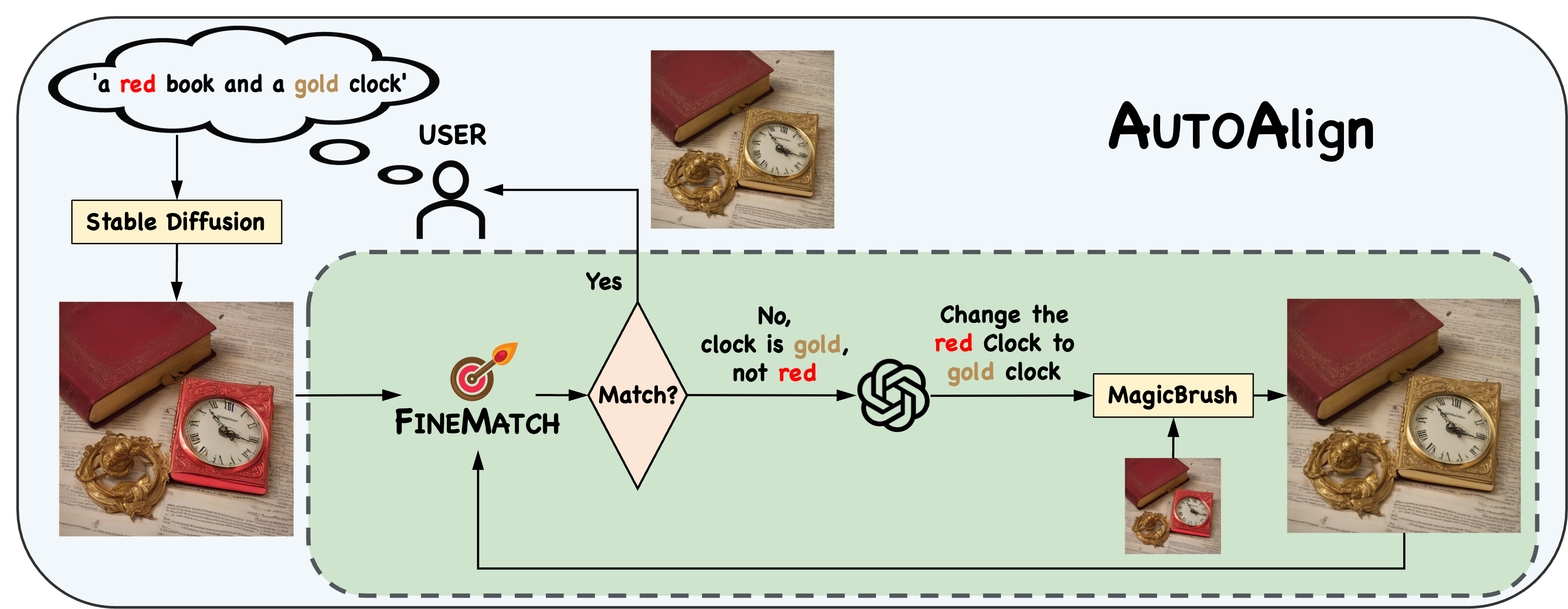}
  \caption{Pipeline of the \AT system: The modules within the system work collaboratively in a loop process until the edited image achieves satisfactory alignment with the text prompt.}
  \label{fig:main_agent}
  \vspace{-5mm}
\end{figure}

The architecture of \AT is shown in Figure~\ref{fig:main_agent}, as we can see from the diagram that the system is designed to effectively reduce the hallucination for T2I generation. More cases are shown in Appendix 6.

\vspace{-1mm}
\section{Limitation and Future Work}
\vspace{-1mm}
This study has a few limitations that present opportunities for further research. First, we provide only one possible correction for each mismatched aspect in the captions. Nevertheless, each mismatched aspect can have several viable corrections. For instance, if an image depicting a woman with blond hair, while the caption is "a woman holding a golden flower", then the correction could be "a golden flower" -> "no golden flower", or it can be "holding"->" with" and "a golden flower"->"blond hair". Recognizing the potential for multiple corrections, we treat the task of mismatched aspect correction as one of open-ended generation. We employ the BERT Score to evaluate the semantic similarity for models' generated corrections and the ground truth. In addition, introducing human evaluation to check if the generated corrections address the mismatch between the image and caption helps evaluate the quality of the predicted corrections of models. Second, finetuning VLMs directly with \NAME has yet to yield results on par with human performance. In the future, we can explore designing better instruction following data for VLMs using \NAME, such as introducing aspect graphs in the text prompt.
\vspace{-4mm}
\section{Conclusion}
\vspace{-2mm}
In this paper, we present \NAME, a novel benchmark designed to address aspect-based, fine-grained mismatches between text and images. We design an aspect graph parsing and node-replacing method combined with human annotation to effectively reduce the artifact biases in the GPT-synthesised data. We also present comprehensive experiments on current state-of-the-art VLMs and found that \NAME can help enhance the ability of the models to perform detailed analyses of text and image mismatches. In addition, we evaluate the current black-box models with strong multimodal in-context-learning capability and find that these models are not skilled at addressing fine-grained mismatches between text and images. With \NAME, we build a text-to-image generation hallucination detection and correction system, and the system can effectively reduce the hallucination for T2I generation. We believe our efforts will benefit real-world applications involving text and image compositional analysis and generation.

\clearpage  % TODO REVIEW/FINAL: This \clearpage needs to be removed from both review and camera-ready versions.

\bibliographystyle{splncs04}
\bibliography{main}
\clearpage
\setcounter{section}{0}
\section{Explanations and Examples for Mismatch Aspects in \NAME}
\label{sec:exp}
\begin{figure*}[htbp]
  \centering
  \includegraphics[width=0.9\textwidth]{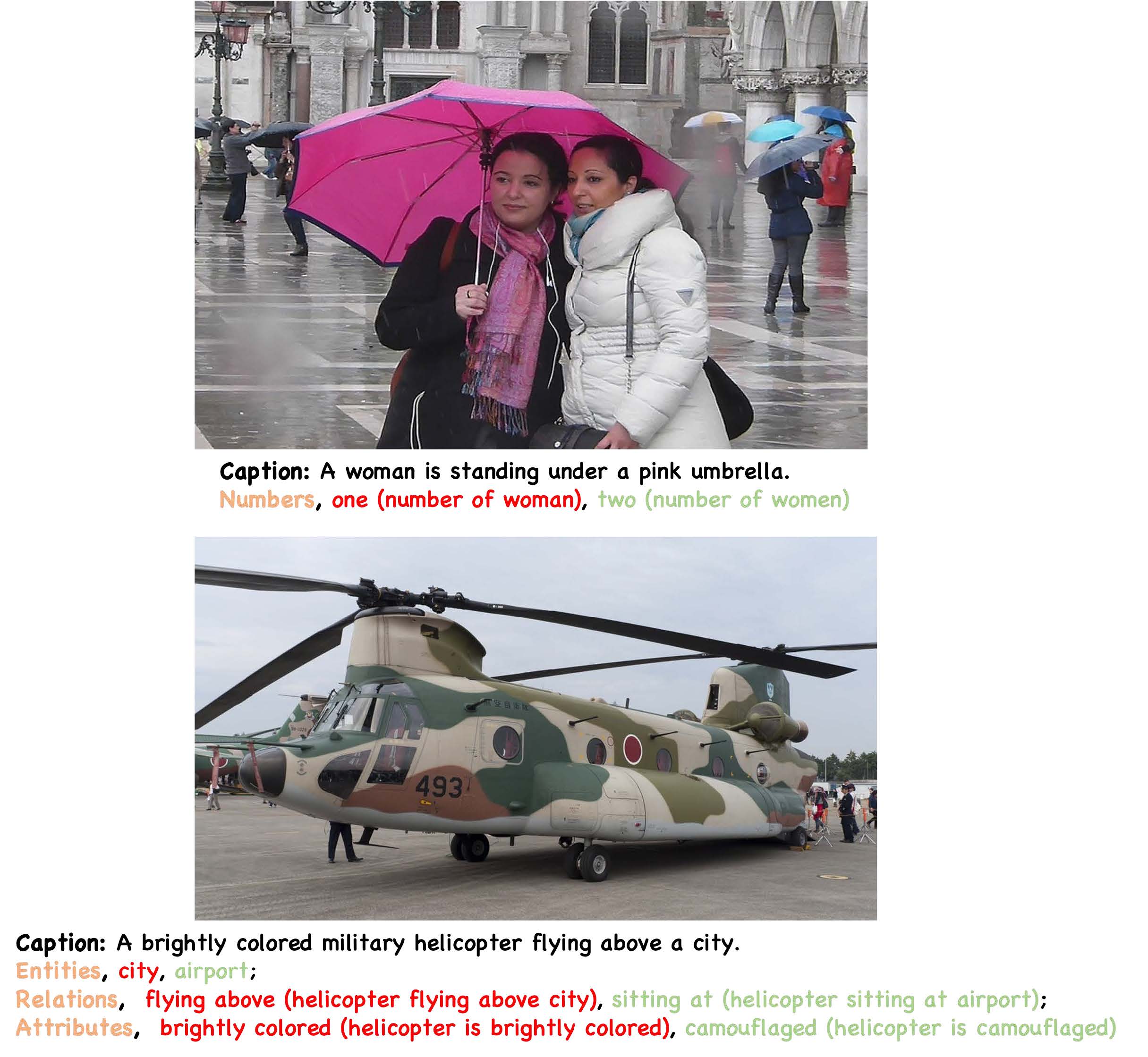}
  \caption{Illustration of the four mismatch aspects in \NAME.}
  \label{fig:case1}
\end{figure*}

We present the definition of the four aspects in \NAME.
\textbf{Entities} are objects within an image that can be recognized by VLMs. In \NAME, we define the entities using the same concepts in Flickr 30K Entities \cite{flickrentitiesijcv}.
The terms \textbf{Attributes} and \textbf{Relations} follow the definitions provided by the Visual Genome (VG) Attribution/Relation framework \cite{krishna2017visual}. However, the categories are limited in VG, while in \NAME the attributes and relations are open-set. \textbf{Numbers} refers to the count of entities in the image. We also provide the examples for each aspect in Figure \ref{fig:case1}.

\section{Statistical Results for \NAME}
\label{sec:qquality}
\begin{figure*}[htbp]
  \centering
  \includegraphics[width=\textwidth]{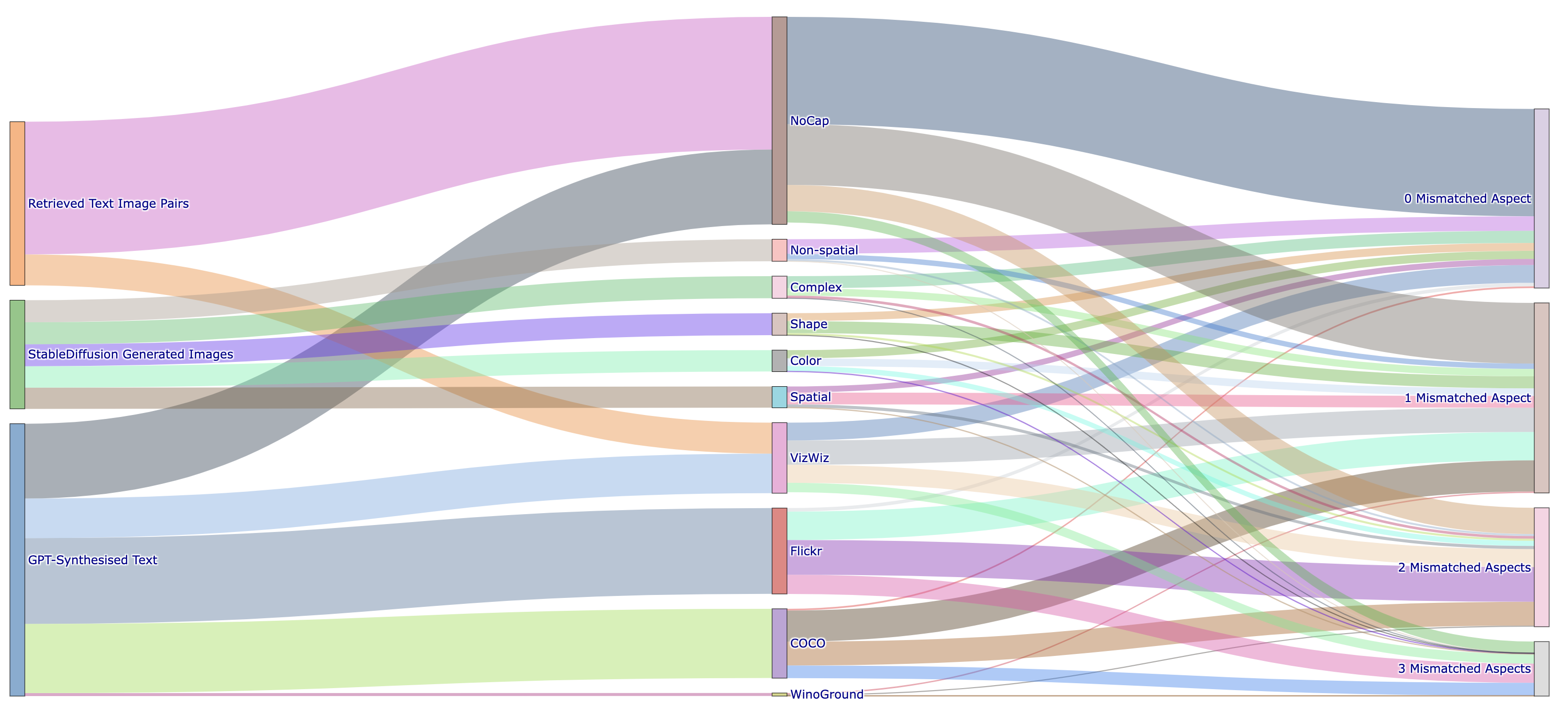}
  \caption{Distribution of demonstrations in \NAME across source, domain, and the number of mismatched aspects in the captions.}
  \label{fig:sankey}
\end{figure*}

\begin{figure*}[htbp]
  \centering
  \includegraphics[width=0.8\textwidth]{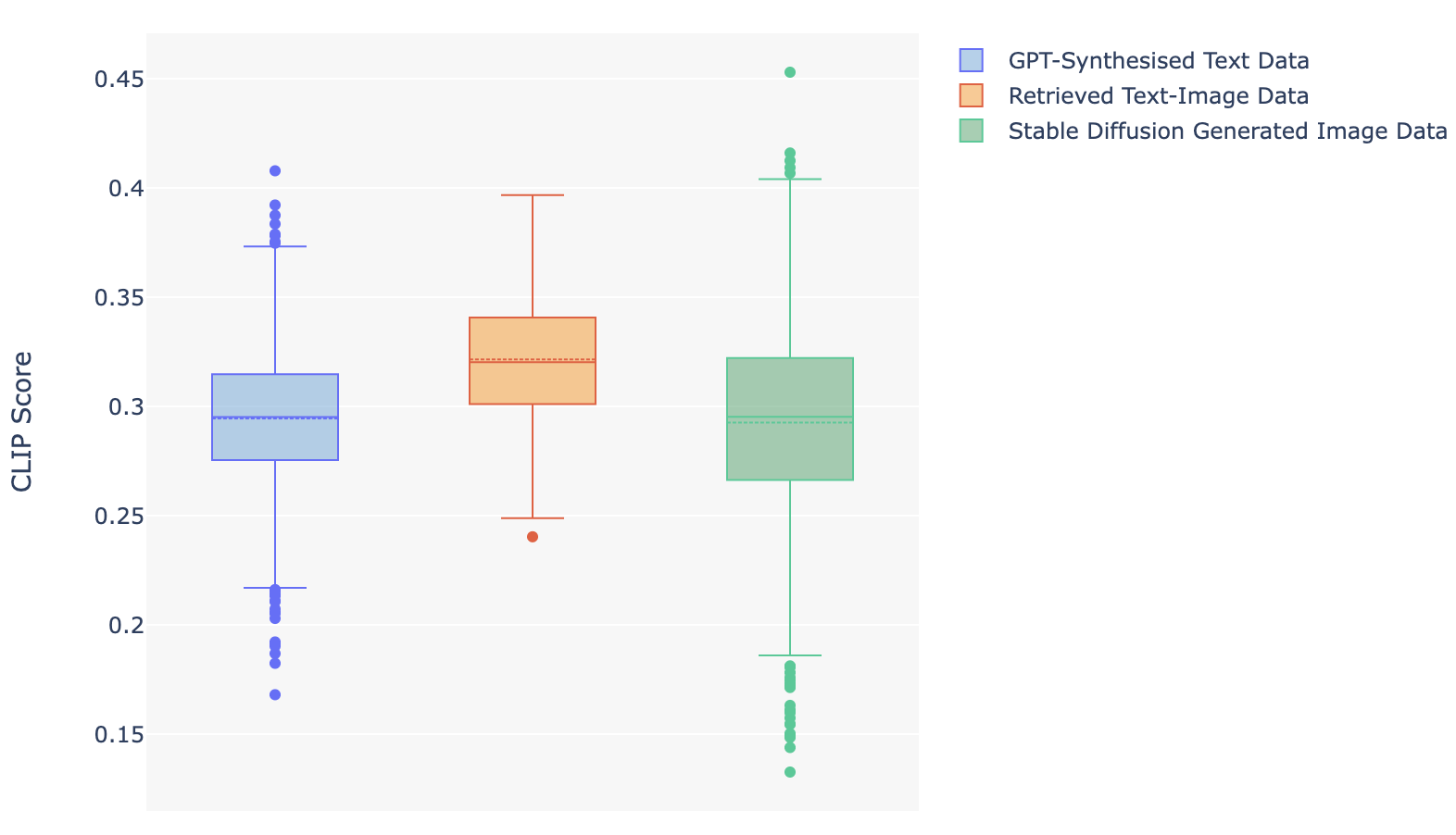}
  \caption{CLIP score distribution across different data sources in \NAME.}
  \label{fig:clip_score}
\end{figure*}

We provide extended statistical results for the quantity and quality analysis across all data sources within \NAME.
Figure \ref{fig:sankey} illustrates the weights and relations distribution among the data source, data domain, and the number of mismatched aspects within the captions. Additionally, Figure \ref{fig:clip_score} shows the distribution of the CLIP scores between the captions and the images in \NAME. Analysis from Figure \ref{fig:clip_score} indicates that the average CLIP score across the three data sources ranges between 0.3 and 0.35. This suggests a comparatively high similarity level between the mismatched captions and the images, making it challenging for models to discern discrepancies between the captions and images.
\label{sec:rationale}

\section{Evaluation Framework}
In this study, the BERT Score is computed using the \texttt{bert-base-uncased} model weights, with the threshold $T$ set as $0.55$ for these conditions. Below, we present the pseudo-code for ITM-IoU. The notation for all variables is consistent with that outlined in Section \ref{sec:eval}.
\begin{algorithm}[htbp]
\caption{ITM-IoU} 
\begin{algorithmic}[1]
\REQUIRE The predicted aspect tuples/triplets $P_i=\{c_j,p_j,o_j\}_{j=1}^M$  over the test set and the ground truth $G_i=\{c_j^\prime,p_j^\prime,o_j^\prime\}_{j=1}^{M^\prime}$ , $i\in |\mathcal{D}|$.
\ENSURE The average ITM-IoU score.
\STATE $Score_{ITM-IoU} \gets \{\}$
\FOR {each $P_i,G_i$}
\STATE $Scores \gets \{\}$

\STATE {$\mathit{Score_{\mathit{Aspect}}}_i \gets \{\}$}
\FOR {$j = 0$ \textbf{to} $M^\prime$}
\STATE {$Score \gets 0$}
\IF {$c_j = c_j^\prime$}
    \STATE $Score \gets W_{Ca}$
\ENDIF

\IF{Mismatch Correction}
\STATE $Score \gets Score + W_{De} \times \mathit{Score_{\mathit{D}}}_j + W_{Co} \times \mathit{Score_{\mathit{C}}}_j$

\ELSE
\STATE {$Score \gets Score + W_{Co} \times \mathit{Score_{\mathit{C}}}_j$}
\ENDIF
\STATE {$\mathit{Score_{\mathit{Aspect}}}_i \gets \mathit{Score_{\mathit{Aspect}}}_i \cup Score$}
\ENDFOR
\IF {$\max( \mathit{Score_{\mathit{Aspect}}}_i) \geq T$}
\STATE {$Scores \gets Scores \cup \max( \mathit{Score_{\mathit{Aspect}}}_i)$}
\STATE \text{Calculate $P_i \cap G_i$}
\ELSE
\STATE {$Scores \gets Scores \cup 0$}
\ENDIF
\STATE \text{Calculate $P_i \cup G_i$}
\STATE $Score_{ITM-IoU} \gets \text{mean}(Scores) \times \frac{P_i \cap G_i}{P_i \cup G_i}$
\ENDFOR
\STATE \textbf{return} $\text{mean}(Score_{ITM-IoU})$
\end{algorithmic}
\label{algo:metrics}
\end{algorithm}

\section{Prompts for Aspect Graph Parsing and Node Replacing for GPT-4V}
\begin{figure}[htbp]
  \centering
  \includegraphics[width=0.9\linewidth]{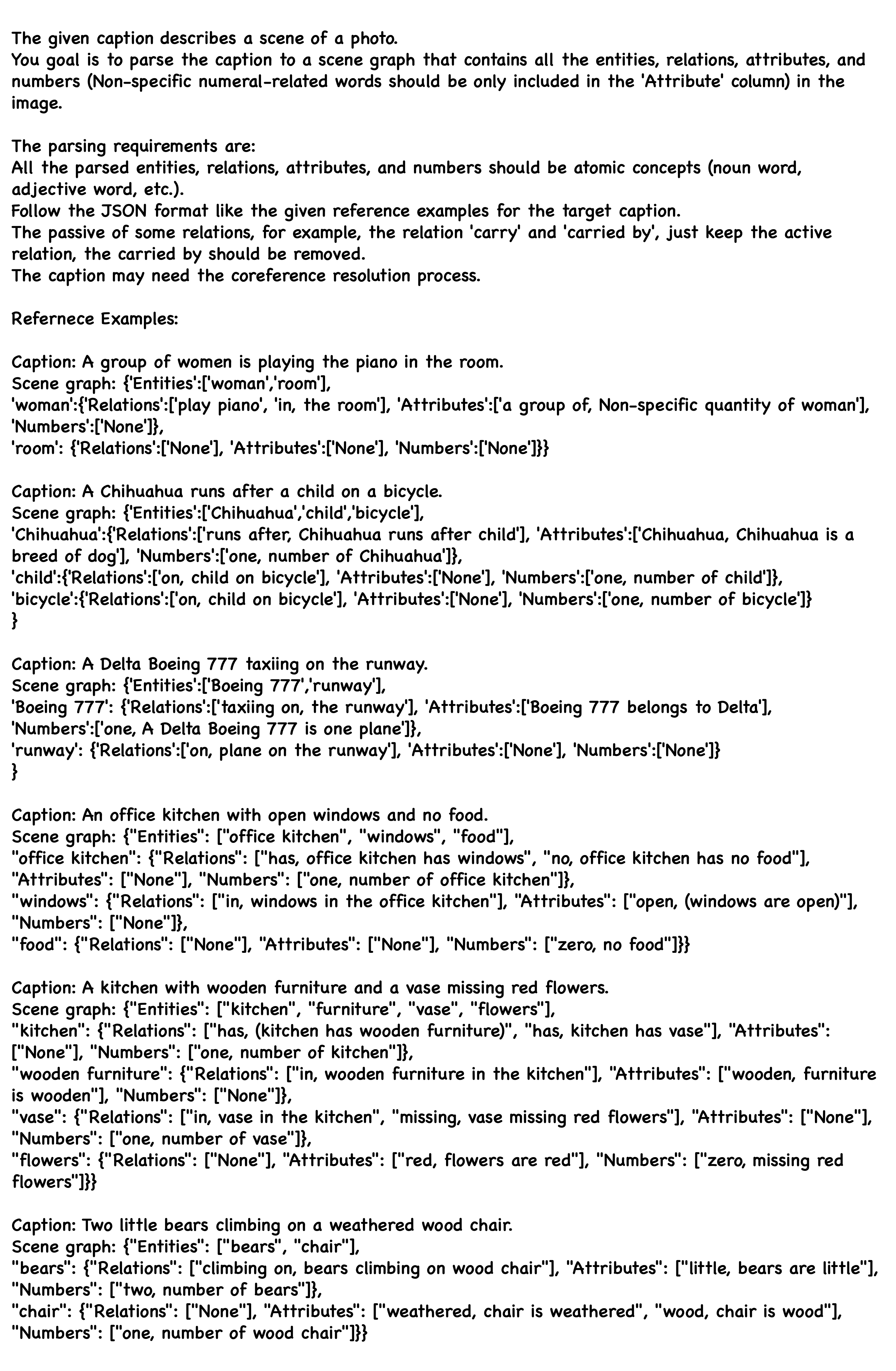}
  \caption{Prompts and context examples for aspect graph parsing with GPT-4.}
  \label{fig:qp}
\end{figure}

\begin{figure}[htbp]
  \centering
  \includegraphics[width=0.9\linewidth]{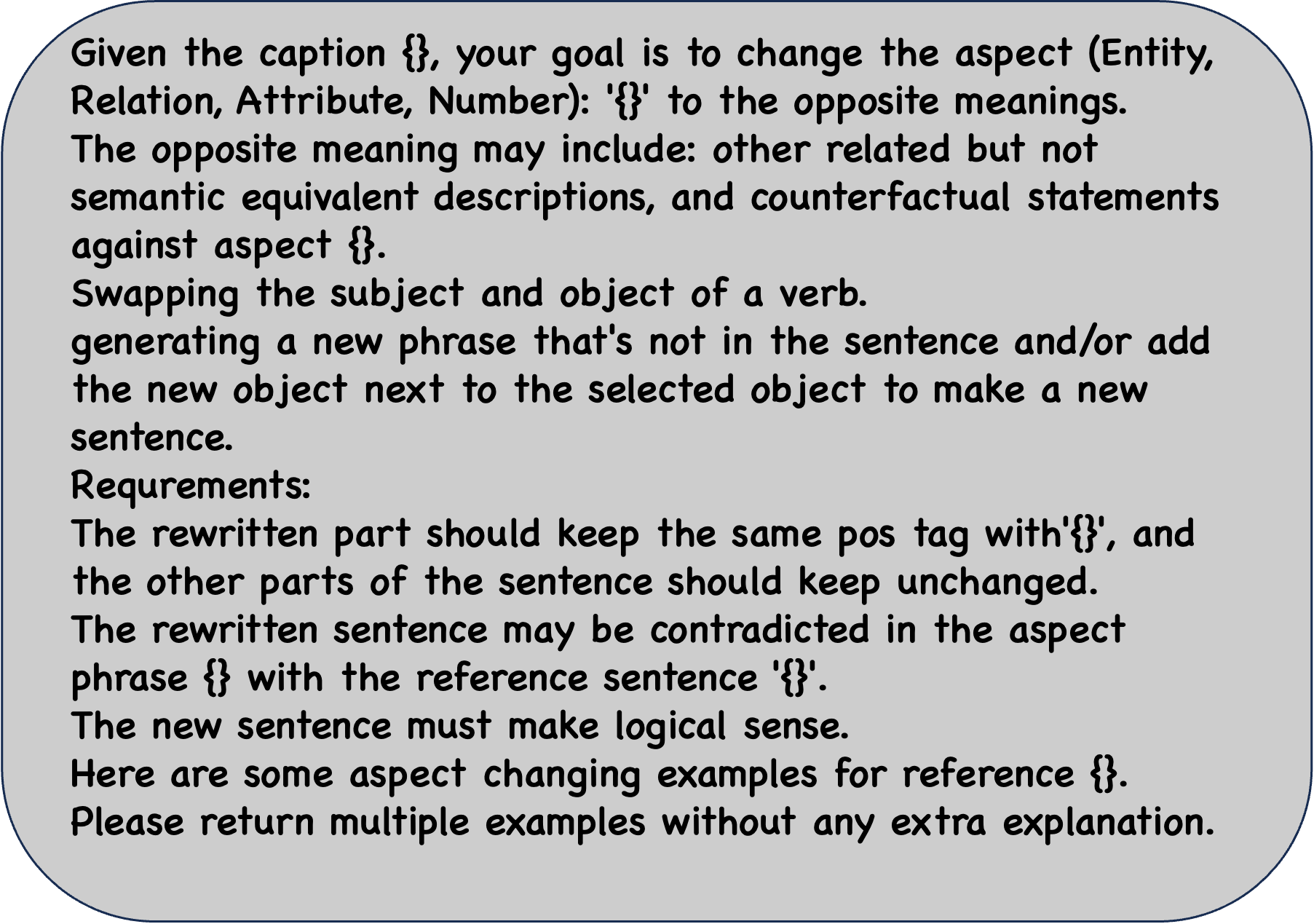}
  \caption{Prompts for node replacement with GPT-4. $\{\}$ in the prompt will be replaced with the initial caption, the node to be replaced $\times 4$, the references, and the context examples.}
  \label{fig:noder}
\end{figure}

Figures \ref{fig:qp} and \ref{fig:noder} illustrate the instructions and context examples for aspect graph parsing and node replacing for GPT-4V, respectively.

\section{Details for Experiments}
\label{sec:fexample}
\subsection{Details for Supervised Learning Experiments}
We conduct supervised learning experiments on \NAME under the visual instruction tuning settings, where we design the instructions to train models for fine-grained image-text mismatch detection and correction. In the experiments, we set the batch size $\in \{128, 256, 512\}$, the learning rate $\in \{1e-5, 3e-5, 5e-5\}$, and we train each model for 3 epochs. We maintain all other hyperparameters at their default values as specified in the official code base. All the models are finetuned on 8 Nvidia A100 GPUs. The predictions of each model are post-processed for calculating  ITM-IoU.

\subsection{Prompts for In-Context Learning Experiments}
\begin{figure}[h!]
  \centering
  \includegraphics[width=\linewidth]{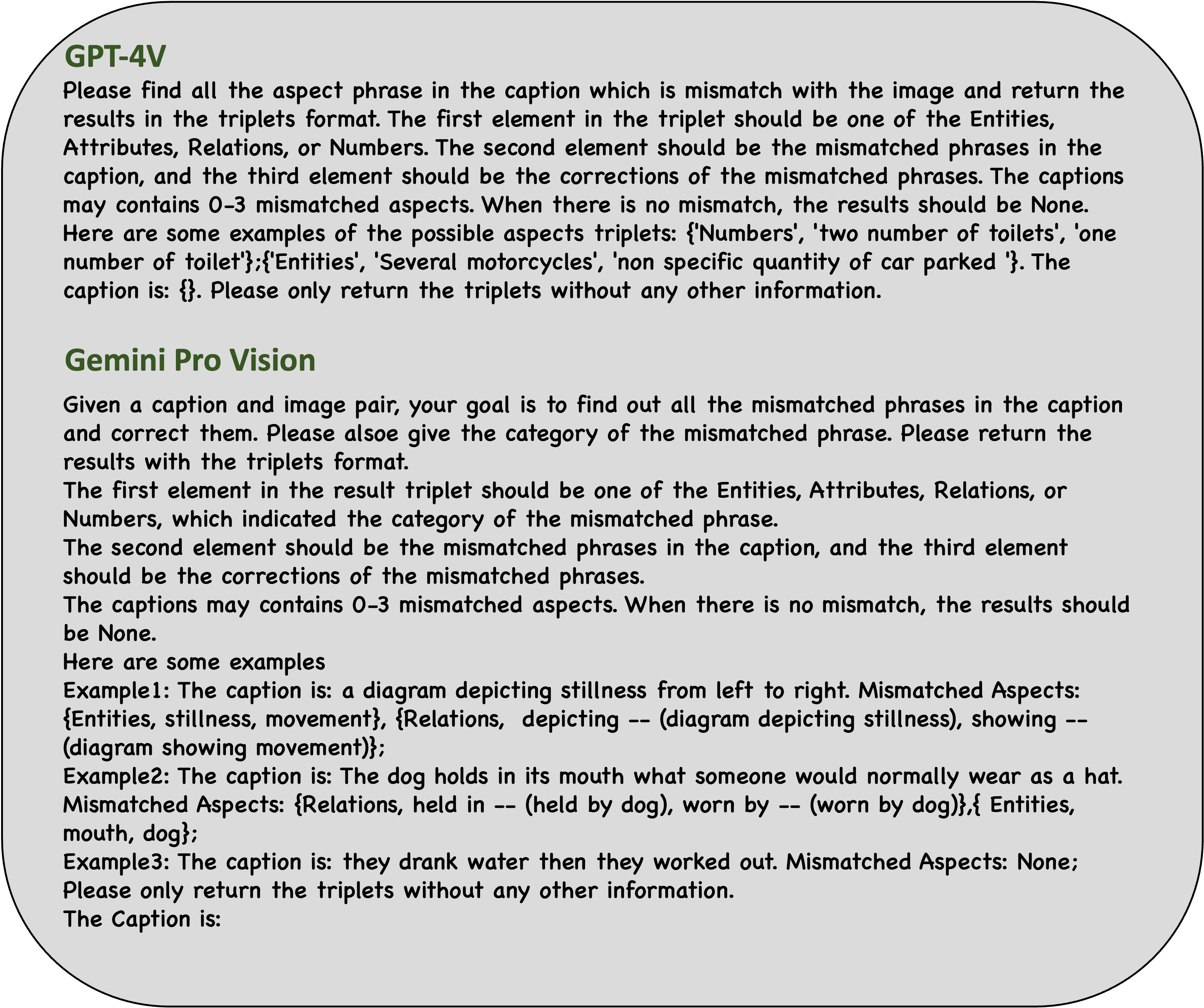}
  \caption{Prompts for the in-context learning experiments with GPT-4V and Gemini Pro Vision.}
  \label{fig:icl}
\end{figure}

We present the prompts for both GPT-4V and Gemini Pro Vison in Figure \ref{fig:icl} for the in-context learning experiments. For each other white-box model, we provide 6 context examples randomly sampled from the training set. 

\section{Examples for \AT}
\label{sec:autocorrect}
\begin{figure}[h!]
  \centering
  \includegraphics[width=\linewidth]{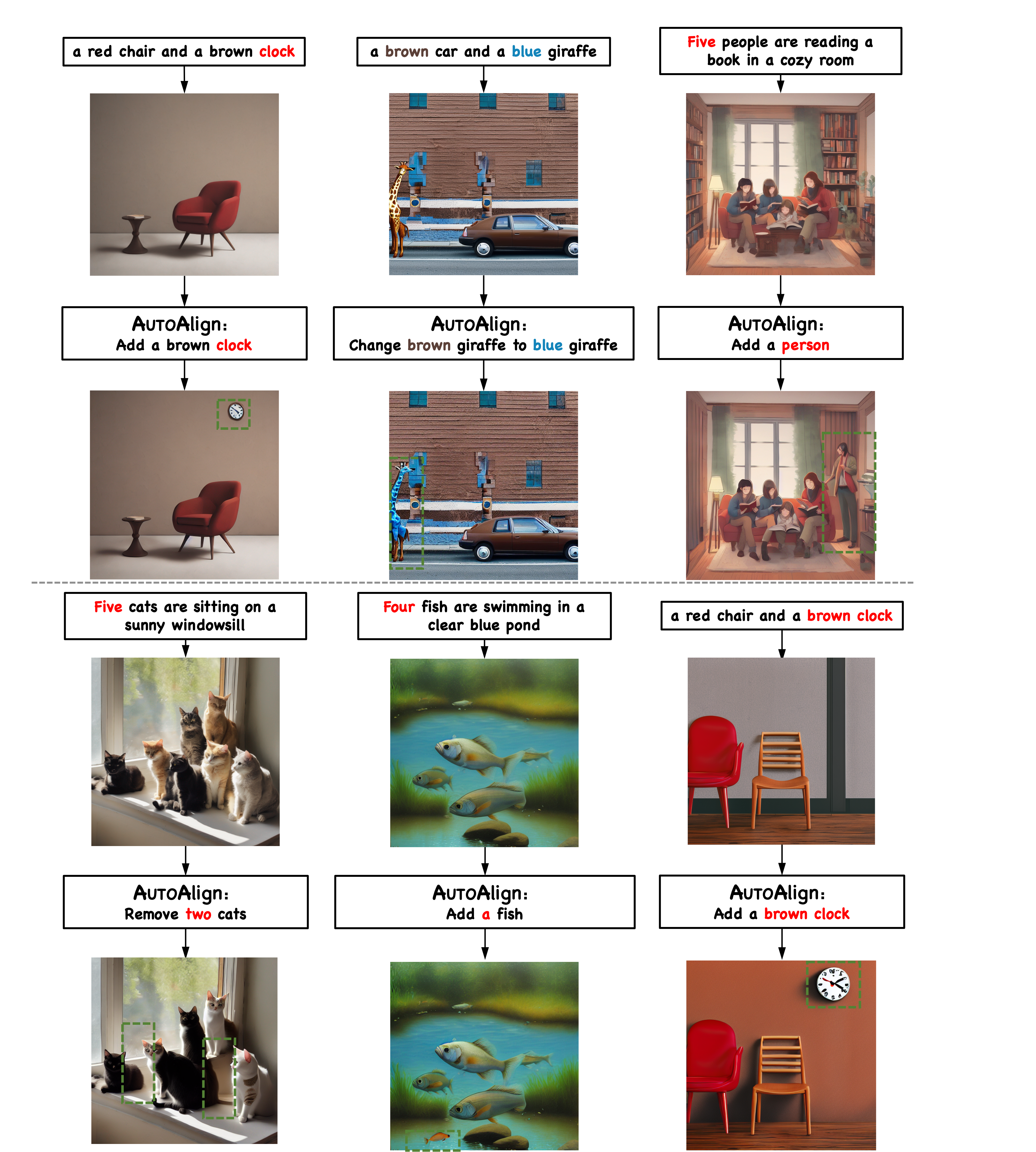}
  \caption{Examples demonstrating how \AT helps reduce the hallucination for T2I generation.}
  \label{fig:cexamples1}
\end{figure}

\begin{figure}[h!]
  \centering
  \includegraphics[width=\linewidth]{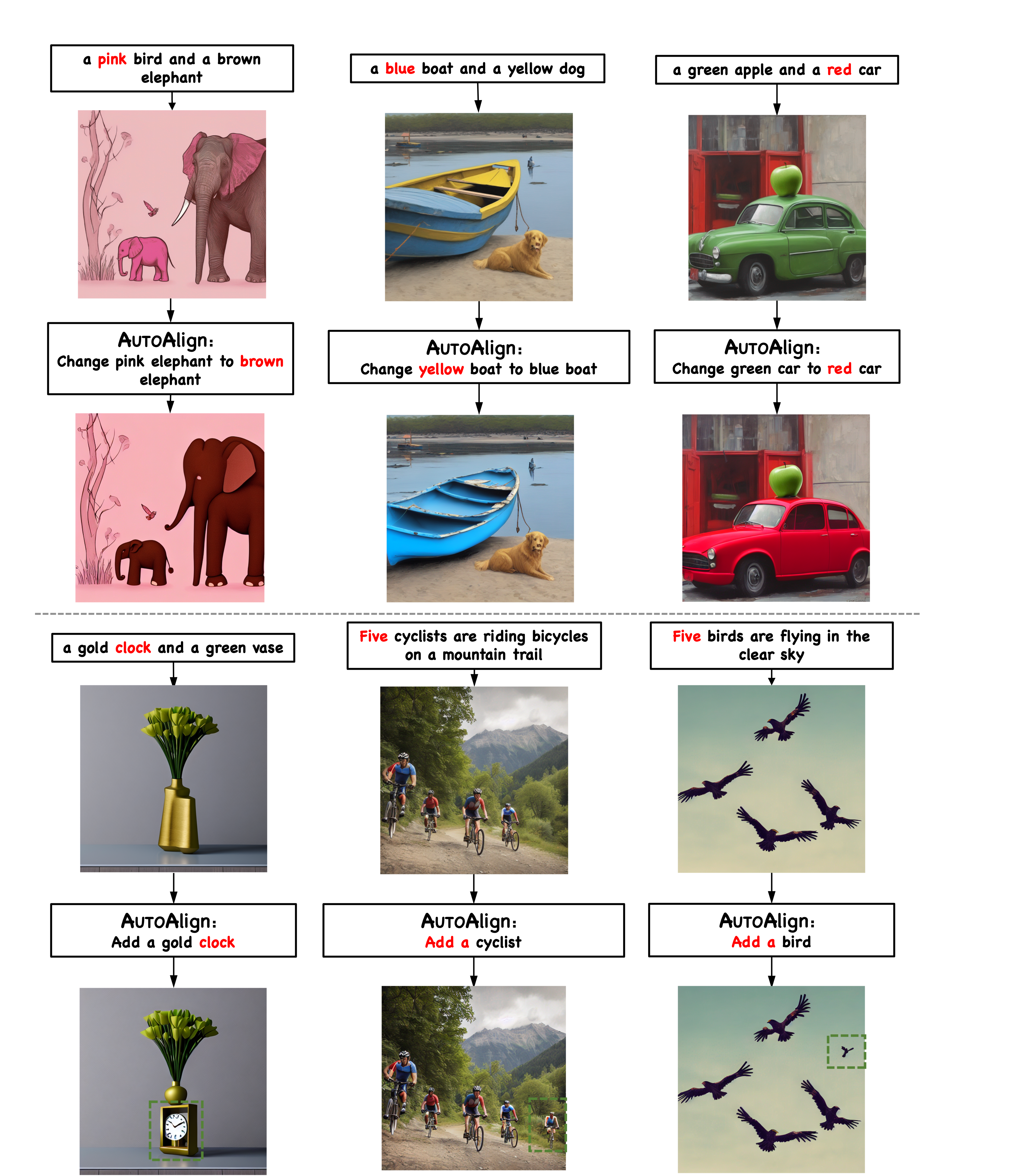}
  \caption{Examples demonstrating how \AT helps reduce the hallucination for T2I generation.}
  \label{fig:cexamples2}
\end{figure}

\begin{figure}[h!]
  \centering
  \includegraphics[width=\linewidth]{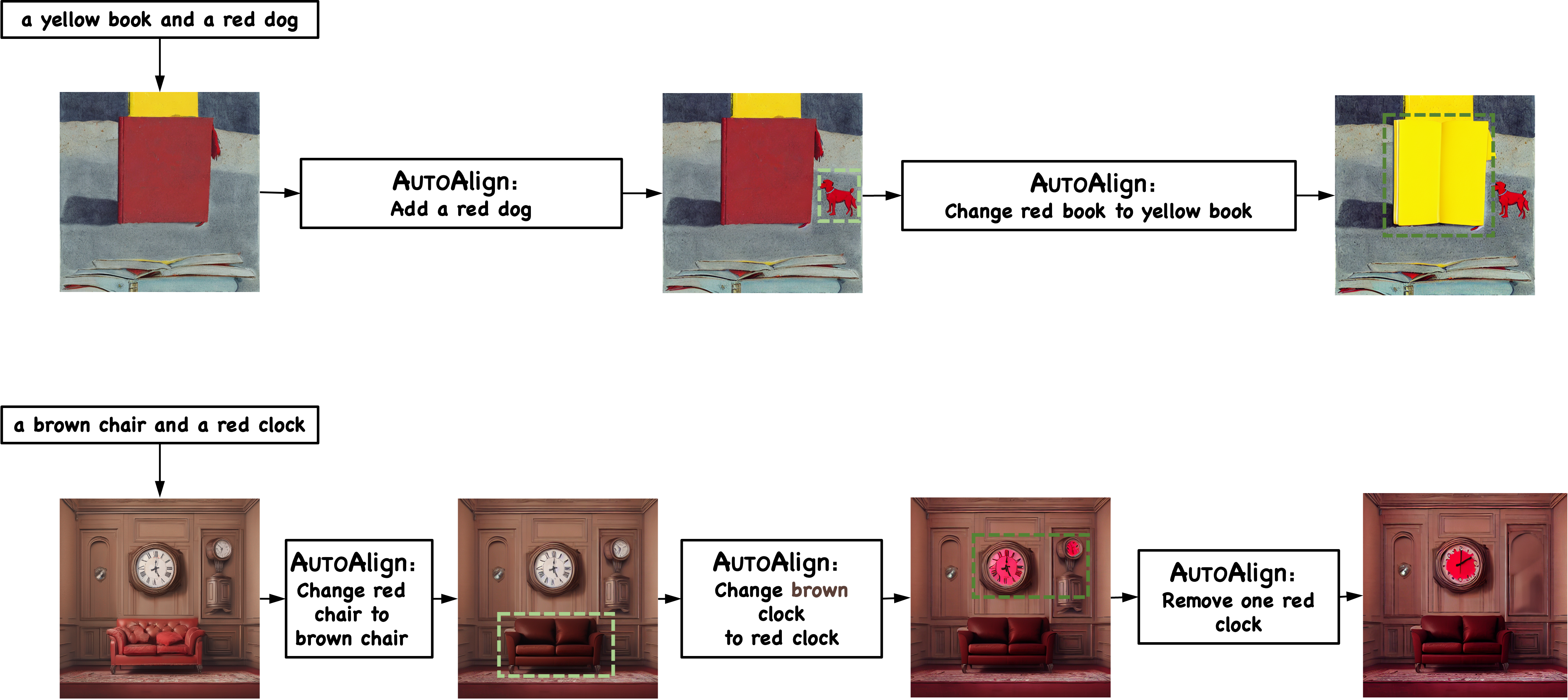}
  \caption{Examples for multi-turn mismatch detection and correction with \NAME .}
  \label{fig:cexamples3}
\end{figure}

\begin{figure}[h!]
  \centering
  \includegraphics[width=0.9\linewidth]{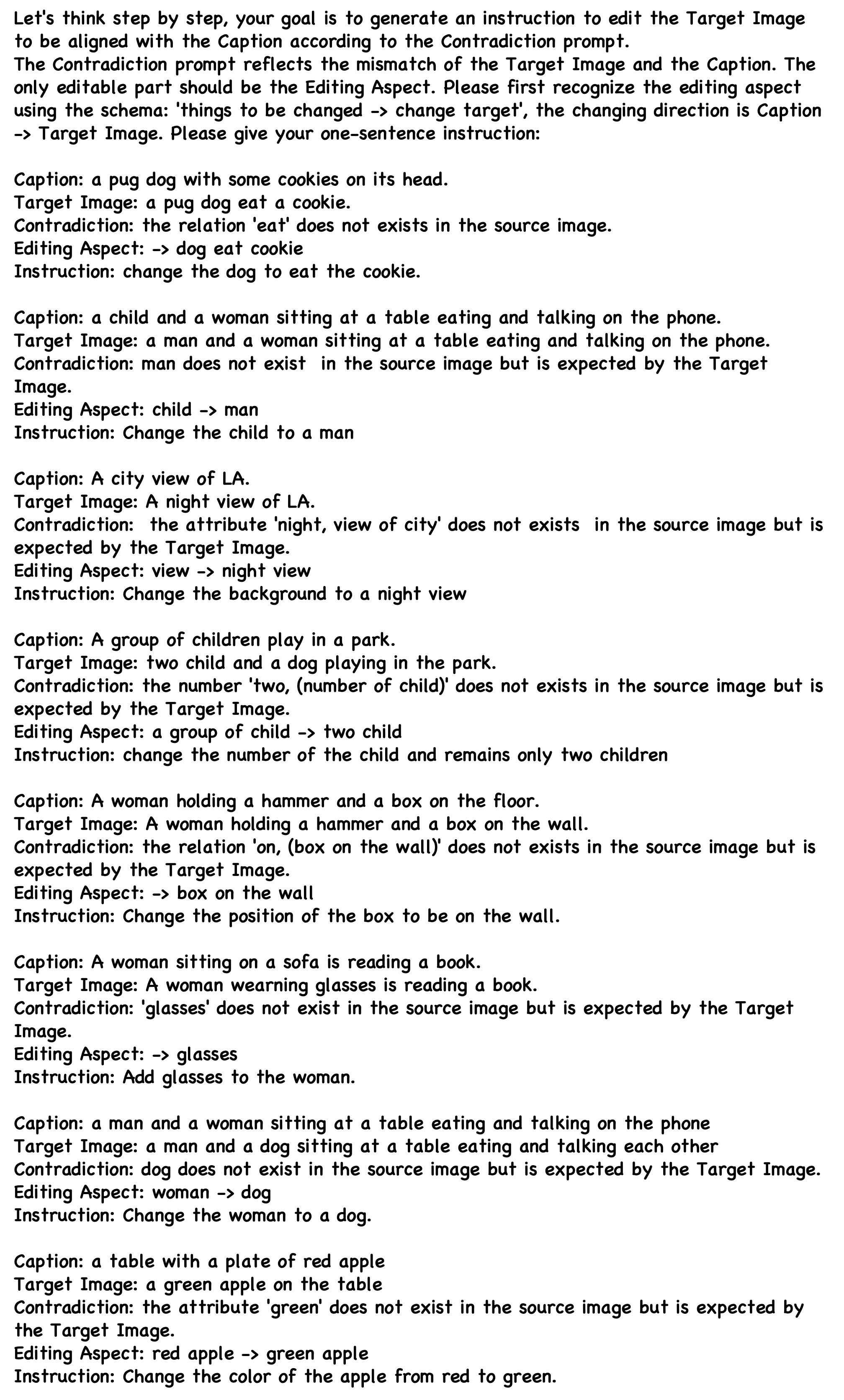}
  \caption{Prompts and context examples for image editing instruction generation with GPT-4.}
  \label{fig:prompte}
\end{figure}

In this section, we provide examples to illustrate that \NAME can help reduce the hallucination for T2I generation. Figure \ref{fig:cexamples1} and Figure \ref{fig:cexamples2} show examples of single-turn mismatch detection and correction. Figure \ref{fig:cexamples3} shows the examples for multi-turn mismatch detection and correction. We also provide the image editing generation prompt for GPT-4V in Figure \ref{fig:prompte}.


\section{Human Annotation}
\label{sec:ha}
\begin{figure*}[h!]
  \centering
  \includegraphics[width=\textwidth]{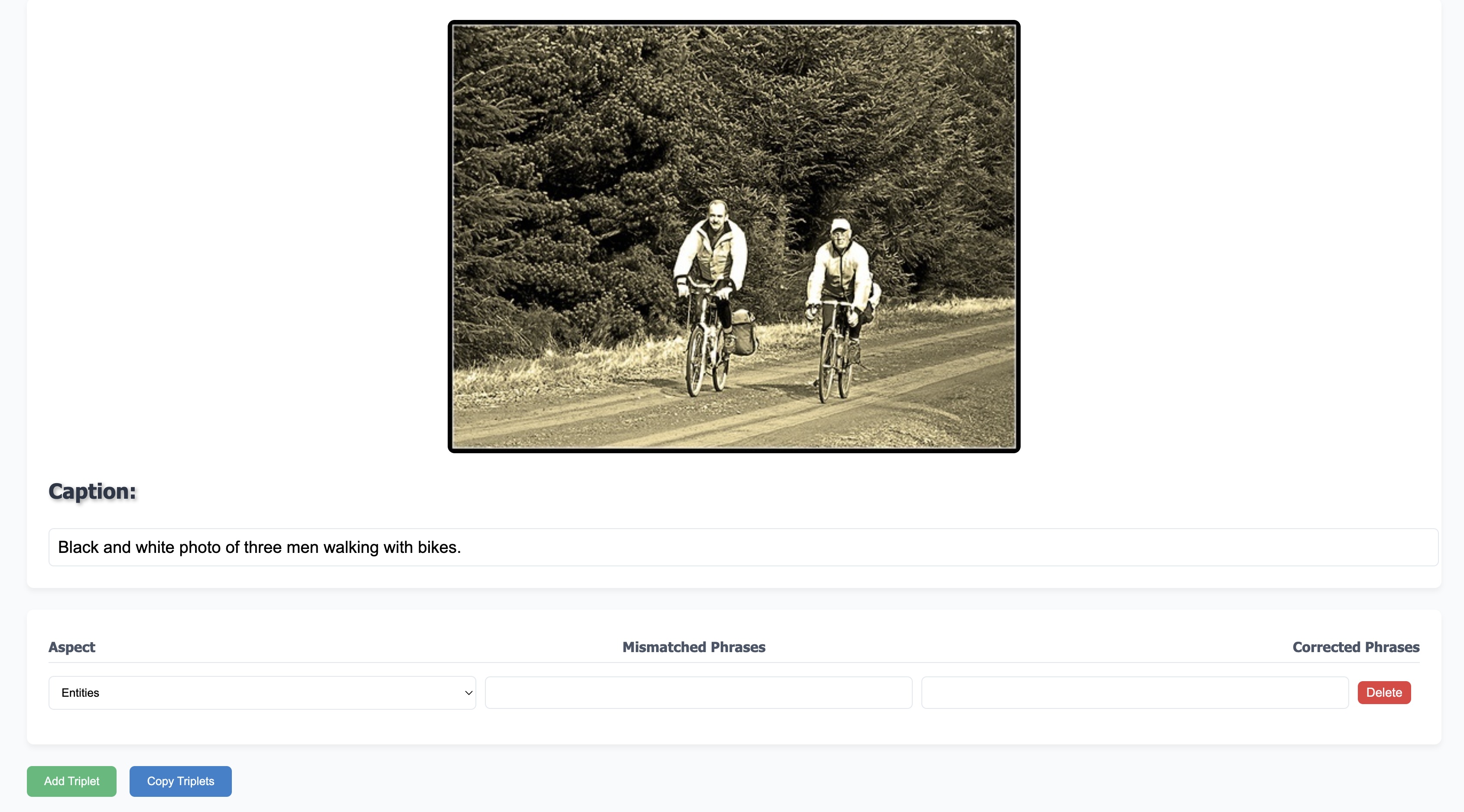}
  \caption{The human annotation interface for \NAME.}
  \label{fig:HA}
\end{figure*}

We design an annotation interface for human experts to annotate the mismatched aspects between the images and the captions. The interface is shown in Figure \ref{fig:HA}. The entire annotation process is conducted on the LabelBox platform.
%Any named entity needs to be explained or any redundant phrase that are irrelevant to the image contents needs to be removed? please give me a yes or no, and parse the target sentence again.

\end{document}